\documentclass[11pt]{article}

\usepackage{epsfig,amsmath,latexsym,amssymb}
\usepackage{graphicx}
\usepackage{lscape}
\usepackage{picture, eso-pic, tikz} 
\usepackage{dsfont}
\usepackage{listings}
\usepackage{changes}
\usepackage{hyperref}
\usepackage{bbding}
\usepackage{tikz-cd}
\usepackage{mathtools}

\renewcommand{\baselinestretch}{1.1}
\def\R{{\mathbb R}}  
\def\N{{\mathbb N}}  
\def\E{{\mathbb E}}  
\def\Beweis{\footnotesize}

\newcommand{\Remm}[1]{}
\newtheorem{theo}{Theorem}[section]

\newtheorem{prop}[theo]{Proposition}

\newtheorem{cor}[theo]{Corollary}

\newtheorem{model ass}[theo]{Model Assumptions}

\newtheorem{example}[theo]{Example}

\newtheorem{rem}[theo]{Remark}
\newtheorem{rems}[theo]{Remarks}

\def\EndProof{\hfill {\scriptsize $\Box$}}
\def\EndExample{\hfill {\scriptsize $\blacksquare$}}
\def\EndRemark{\hfill {\scriptsize $\blacksquare$}}
\numberwithin{equation}{section}

\definecolor{MyGray}{rgb}{0.92,0.92,0.92}


\definecolor{British racing}{rgb}{0.0, 0.5, 0.0}
\def\bx{\boldsymbol{x}}

\def\bZ{\boldsymbol{Z}}
\def\bY{\boldsymbol{Y}}

\def\bB{\boldsymbol{B}}

\def\bbeta{\boldsymbol{\beta}}

\def\bphi{\boldsymbol{\phi}}
\def\bOne{\boldsymbol{1}}
\def\b0{\boldsymbol{0}}

\usepackage{todonotes}
\newcommand{\Comments}{1}
\newcommand{\mynote}[2]{\ifnum\Comments=1\textcolor{#1}{#2}\fi}
\newcommand{\mytodo}[2]{\ifnum\Comments=1%
  \todo[linecolor=#1!80!black,backgroundcolor=#1,bordercolor=#1!80!black]{#2}\fi}

\ifnum\Comments=1               
\fi

\ifnum\Comments=1               
\fi

\ifnum\Comments=1               
\fi

\begin{document}

\author{Michael Mayer\footnote{Actuarial Department, La Mobilière, Bern, michael.mayer@mobiliar.ch} \and Mario V.~W{\"u}thrich\footnote{\normalfont Department of Mathematics, ETH Zurich, mario.wuethrich@math.ethz.ch}}

\date{\today\\~}
\title{{\sc Shapley Values: Paired-Sampling Approximations}\\~}
\maketitle

\begin{abstract}
  \noindent
 Originally introduced in cooperative game theory, Shapley values have become a very popular tool to explain machine learning predictions. Based on Shapley's fairness axioms, every input (feature component) gets a credit how it contributes to an output (prediction). These credits are then used to explain the prediction. The only limitation in computing the Shapley values (credits) for many different predictions is of computational nature.  There are two popular sampling approximations, sampling KernelSHAP and sampling PermutationSHAP. Our first novel contributions are asymptotic normality results for these sampling approximations. Next, we show that the paired-sampling approaches provide exact results in case of interactions being of maximal order two. Furthermore,  the paired-sampling PermutationSHAP  possesses the additive recovery property, whereas its kernel counterpart does not.

~ 
 
\bigskip

\end{abstract}

~

\section{Introduction}

Shapley \cite{Shapley} values have become one of the most popular tools in explaining predictions of machine learning and statistical models, but they are also used in many other contexts, e.g., in actuarial science for risk allocation in variable annuities, see Godin et al.~\cite{Godin}, or the construction of actuarial rating systems, see Vallarino \cite{Vallarino}. Shapley values have originally been introduced in cooperative game theory. They are concerned about a fair allocation of gains and losses (total payoffs) to the individual players of a cooperative game.

Assume there are $q\ge 2$ players and set ${\cal Q}=\{1,\ldots, q\}$ for the grand coalition. Assume there is a {\it value function}
\begin{equation*}
\nu: {\cal C} ~\mapsto~ \nu({\cal C})\in\R,
\end{equation*}
that measures the contributions of coalitions ${\cal C} \subset {\cal Q}$ to the total payoff $\nu({\cal Q})$ of the cooperative game. Shapley \cite{Shapley} postulated the following axioms to be desirable properties of a fair allocation $(\phi_j)_{j=1}^q=(\phi^{(\nu)}_j)_{j=1}^q$ of the total payoff $\nu({\cal Q})$ to the $q$ players:
\begin{itemize}
	\item[(A1)] {\em Efficiency:} $\nu({\cal Q}) = \sum_{j = 0}^q \phi_j$, with $\phi_0$ denoting the non-distributed payoff.
	\item[(A2)] {\em Symmetry:} If $\nu({\cal C} \cup \{j\}) = \nu({\cal C} \cup \{k\})$ for every ${\cal C} \subset {\cal Q} \setminus\{j, k\}$, then $\phi_j = \phi_k$. 
	\item[(A3)] {\em Dummy player:} If $\nu({\cal C} \cup \{j\}) = \nu({\cal C})$ for every ${\cal C} \subseteq {\cal Q} \setminus\{j\}$, then $\phi_j = 0$.
	\item[(A4)] {\em Linearity:} For two cooperative games with value functions $\nu$ and $\mu$, there is linearity $\phi_j^{(\nu + \mu)} = \phi_j^{(\nu)} + \phi_j^{(\mu)}$ and $\phi_j^{(\alpha \nu)} = \alpha \phi_j^{(\nu)}$ for all $j \in {\cal Q}$ and $\alpha \in \R$.
\end{itemize}
Shapley \cite{Shapley} proved that for a given value function $\nu$, there is exactly one solution $(\phi_j)_{j=0}^q$ satisfying these four axioms, and it is given by
\begin{equation}\label{Shapely decomposition}
	\phi_j = \frac{1}{q} \sum_{{\cal C} \subseteq {\cal Q} \setminus\{j\}}
	{q-1 \choose |{\cal C}|}^{-1}
	\left(\nu({\cal C} \cup \{j\}) - \nu({\cal C})\right) \qquad \text{ for all $j \in {\cal Q}$,}
\end{equation}
and non-distributed payoff $\phi_0 = \nu({\cal Q})-\sum_{j=1}^q \phi_j$.

\medskip

W.l.o.g.~we may and will assume that $\phi_0=\nu(\emptyset)=0$, because otherwise we simply consider the translated value function $\nu_0({\cal C})=\nu({\cal C})-\nu(\emptyset)$,
and we receive exactly the same solution \eqref{Shapely decomposition}
of the Shapley values $(\phi_j)_{j=1}^q$ for the translated value function $\nu_0$.
In fact, this is justified by the linearity axiom. Decomposing the value function $\nu({\cal C})=\nu_0({\cal C})+\phi_0$, the last term is a constant value function $\phi_0\equiv \phi_0({\cal C})$ that makes every player to a dummy player in this second game, hence there is no distribution to individual players from this second constant value function.
	
\medskip

These Shapley values $(\phi_j)_{j=1}^q$ have become a standard model-agnostic tool in explaining machine learning predictions, where the role of the players is taken over by the feature components, and the value function represents the prediction made by the feature. This is known as the SHapley Additive exPlanation (SHAP); see Lundberg--Lee \cite{lundberg2017}.

\medskip

There is a critical point though, namely, the computation of the Shapley values \eqref{Shapely decomposition} scales badly in the dimension $q$ of the feature space -- size of the grand coalition ${\cal Q}$ -- because it considers all possible coalitions ${\cal C}$ of ${\cal Q}$, and this can make the SHAP computation \eqref{Shapely decomposition} prohibitive; there are $2^q$ coalitions ${\cal C}\subseteq {\cal Q}$. For this reason, alternative equivalent representations of \eqref{Shapely decomposition} have been derived, and simulation algorithms have been employed to efficiently approximate the Shapley values $(\phi_j)_{j=1}^q$ for a given value function $\nu$. The main goal of this paper is to review these alternative representations and approximations, and we provide some new mathematical results.
There are two main equivalent representations of the Shapley values \eqref{Shapely decomposition}, we briefly introduce them next.

\paragraph{PermutationSHAP} The PermutationSHAP representation
has been presented in Castro et al.~\cite{Castro} and \v{S}trumbelj--Kononenko \cite{Strumbelj2010, Strumbelj2014}.
Denote by $\pi=(\pi_1,\ldots, \pi_q )$ a permutation of the ordered set $(1,\ldots, q)$. Let $\kappa(j) \in {\cal Q}$ be the index with $\pi_{\kappa(j)}=j$, and define the set
\begin{equation}\label{kappa function}
{\cal C}_{\pi, j} = \left\{ \pi_1,\ldots, \pi_{\kappa(j)-1}  \right\}~\subset ~{\cal Q}.
\end{equation}
These are the components of $\pi$ preceding $\pi_{\kappa(j)}=j$; if $\kappa(j)=1$, i.e., if $\pi_1=j$, this is the empty set $\emptyset$. 
The $j$-th Shapley value \eqref{Shapely decomposition} is equivalently obtained by 
\begin{equation}\label{permutation SHAP}
\phi_j = \frac{1}{q!} \sum_{\pi}\nu\left({\cal C}_{\pi, j}\cup \{j\}\right)
-\nu\left({\cal C}_{\pi, j}\right) .
\end{equation}
The main difference between \eqref{Shapely decomposition}
and \eqref{permutation SHAP} is that the permutations $\pi$ of ${\cal Q}$ provide the correct weights (multiplicity) of the coalitions in 
\eqref{Shapely decomposition}.

\paragraph{KernelSHAP} The KernelSHAP representation has been introduced
by Lundberg--Lee \cite[Theorem 2]{lundberg2017}. 
The Shapely values $(\phi_j)_{j=0}^q$ minimize the objective function
\begin{equation}\label{KernelSHAP}
\sum_{\emptyset\neq {\cal C} \subsetneq {\cal Q}}\, \frac{q-1}{{q \choose |{\cal C}|}|{\cal C}|(q-|{\cal C}|)}\, \left(\nu({\cal C}) -\phi_0 - \sum_{j\in{\cal C}}\phi_j\right)^2,
\end{equation}
under the two side constraints $\phi_{0} = \nu(\emptyset)$ and $\phi_{0} + \sum_{j=1}^q\phi_{j} = \nu(\cal Q)$. Again, w.l.o.g.~we may and will assume $\nu(\emptyset)=0$, this allows us to drop $\phi_0=0$ from all considerations done below.

\paragraph{Contributions}
We emphasize that \eqref{Shapely decomposition}, \eqref{permutation SHAP}
and \eqref{KernelSHAP} give three equivalent representations which all suffer the same computational complexity for large $q$. Therefore, sampling versions have been proposed to approximate the (exact) Shapley values.
(1) Our first contribution is to discuss these sampling versions, this discussion essentially follows the results in  \v{S}trumbelj--Kononenko \cite{Strumbelj2010, Strumbelj2014}, Lundberg--Lee \cite{lundberg2017} and Covert--Lee \cite{covert2021}.
(2) Our second contribution is to provide asymptotic normality results for these sampling versions. To the best of our knowledge, these results are new, and they are practically relevant because they indicate the necessary minimal sample size to receive reliable empirical approximations. This is particularly important in view of limited computational budgets, which is a key issue in Shapley value calculations.
(3) Our third contribution is to pick up a statement in the SHAP manual of Lundberg \cite{Lundberg2018B} saying that the sampling version of the paired PermutationSHAP is exact in case of a value function with interaction terms of maximal order 2. We formally prove this result, in fact, one single permutation is sufficient in this case to get the exact Shapley values. Moreover, we state and prove an analogous novel result for the sampling version of the KernelSHAP.
(4) Finally, we provide further results on the additive recovery property which only hold for the paired-sampling PermutationSHAP, but not for its paired kernel counterpart. These latter results suggest to give preference to the PermutationSHAP version.

\medskip

We close this introduction with remarks. First, the subsequent analysis is based on a given value function $\nu$, and we quantify the sampling approximation errors to the (exact) Shapley values using the sampling algorithms being inspired by the PermutationSHAP \eqref{permutation SHAP} and  the KernelSHAP \eqref{KernelSHAP}. There is another stream of literature that discusses how the value function $\nu$ can be selected for a machine learning model; see, e.g., Covert et al.~\cite{covert2020}. 
A machine learning or statistical model $\mu$ gives for every input (feature) $\bx$
an output $\mu(\bx)$, which may form a prediction of a random variable. Shapley values are used to explain this prediction $\mu(\bx)$ in terms of the input $\bx$. This suggests to take a value function $\nu(\cdot)=\nu_{\bx}(\cdot)$ that provides for the grand coalition ${\cal Q}$ the total payoff $\nu({\cal Q})=\nu_{\bx}({\cal Q}):=\mu(\bx)$. The main question now is how to select the value function 
$\nu({\cal C})=\nu_{\bx}({\cal C})$ for coalitions ${\cal C} \subsetneq {\cal Q}$; see Covert et al.~\cite{covert2020}.  Intuitively, $\nu({\cal C})=\nu_{\bx}({\cal C})$ should describe the prediction made if certain feature components of $\bx$ are masked.
An essential point in these discussions is whether one considers conditional or unconditional (interventional) versions of the predictions if certain feature components drop out (are not part of the considered coalition).
We will not enter this discussion here, but work under a fixed given value function $\nu$. However, given that the same technique to estimate the value function is used, our results on equivalence of methods translate one-to-one to practical machine learning applications using either of the two versions of the value function definition.

Second, the theory presented here is model-agnostic, i.e., we assume a fixed given value function $\nu$, but we do not assume that this value function has been obtained by a certain algorithm (model). There are model specific versions, e.g., if $\nu$ is obtained by a decision tree construction, it has a tree structure which can be very beneficial in computing the Shapley values efficiently. A specific example is TreeSHAP of Lundberg et al.~\cite{lundberg2020, lundberg2018}. We will not discuss this in detail here, but we mention that TreeSHAP usually leads to different results compared to the sampling versions of PermutationSHAP and KernelSHAP. A main reason for this difference  is that TreeSHAP is based on a conditional version of the value function definition (this is naturally and efficiently obtained from the tree structure), whereas the other two sampling SHAP algorithms typically consider an interventional version of the value function (because this can be computed more efficiently in case of an absent tree structure).

\bigskip

{\bf Organization.} Section \ref{sec Sampling KernelSHAP} discusses sampling versions of KernelSHAP, it gives the asymptotic normality statements, and it proves the accuracy results in the case of a value function of maximal order two.
Section \ref{sec Sampling PermutationSHAP} focuses on PermutationSHAP, and it essentially proves the equivalent results in this second SHAP version. All technical proofs are given in the appendix. Section \ref{Additive recovery property}
discusses the additive recovery property in more generality, and we prove that it holds for the paired PermutationSHAP, but not the KernelSHAP version.
Finally, in Section \ref{Section Conclude}, we conclude.

\section{Sampling KernelSHAP}
\label{sec Sampling KernelSHAP}
\subsection{KernelSHAP}
We first rewrite the KernelSHAP objective function \eqref{KernelSHAP} and we integrate the side constraint. Throughout we assume $\nu(\emptyset)=0$.
The KernelSHAP optimization problem \eqref{KernelSHAP}
is a weighted least squares problem with weights
(for convenience we normalize to probability weights)
\begin{equation*}
p({\cal C}) =\left(\sum_{\emptyset\neq {\cal C'} \subsetneq {\cal Q}}\frac{q-1}{{q \choose |{\cal C'}|}|{\cal C'}|(q-|{\cal C'}|)}\right)^{-1}\frac{q-1}{{q \choose |{\cal C}|}|{\cal C}|(q-|{\cal C}|)}~>~0.
\end{equation*}
Following Covert--Lee \cite{covert2021}, we introduce a binary notation for the coalitions ${\cal C}$ by rewriting them as $q$-dimensional vectors $\bZ=\bZ_{\cal C}=(Z_j)_{j=1}^q \in \{0,1\}^q$ indicating whether index $j$ is part of the coalition ${\cal C}$, $Z_j=1$, or not, $Z_j=0$. Thus, every coalition ${\cal C}$ can be identified (one-to-one) by a binary vector $\bZ$, and we set, by a slight abuse of notation, $\nu(\bZ)=\nu({\cal C})$. Moreover, we introduce the vector notation $\bphi =(\phi_j)_{j=1}^q \in \R^q$, and we denote the $q$-dimensional zero-vector and one-vector by $\b0\in \R^q$ and $\bOne\in \R^q$, respectively. This allows us to reformulate \eqref{KernelSHAP}
 as, see Covert--Lee \cite[formula (5)]{covert2021}
\begin{equation}\label{KernelSHAP 2}
\underset{\bphi \in \R^q}{\arg\min}\,\E_{\bZ \sim p}
\left[ \left(\nu(\bZ) - \bZ^\top \bphi\right)^2\right]
\qquad \text{ subject to $\bOne^\top\bphi = \nu(\bOne)$.}
\end{equation}
In statistics, minimization problem \eqref{KernelSHAP 2} is called an M-estimation problem with side constraint. We solve it for the side constraint $\phi_q = \nu(\bOne)- \sum_{j=1}^{q-1} \phi_j$, providing us with the unconstraint M-estimation problem
\begin{equation}\label{KernelSHAP 3}
\underset{\widetilde{\bphi} \in \R^{q-1}}{\arg\min}\,\frac{1}{2}\,\E_{\bZ \sim p}
\left[ \left(\nu(\bZ) -Z_q \nu(\bOne)- \left(\widetilde{\bZ}-Z_q\widetilde{\bOne}\right)^\top \widetilde{\bphi}\right)^2\right];
\end{equation}
we generically use the tilde notation for $(q-1)$-dimensional vectors $\widetilde{\bZ}$ by dropping the $q$-th component $Z_q$ from the $q$-dimensional ones $\bZ$. 

We turn this M-estimation into a Z-estimation problem by computing the gradient w.r.t.~$\widetilde{\bphi}$. The optimal solution $\widetilde{\bphi}$ to \eqref{KernelSHAP 3} is found by solving the score equations
\begin{equation}\label{KernelSHAP 4}
\E_{\bZ \sim p}
\left[ \left(\nu(\bZ)-Z_q \nu(\bOne) - \left(\widetilde{\bZ}-Z_q\widetilde{\bOne}\right)^\top \widetilde{\bphi}\right)\left(\widetilde{\bZ}-Z_q\widetilde{\bOne}\right)\right]=\widetilde{\b0}.
\end{equation}
The solution to \eqref{KernelSHAP 4} is given by the $(q-1)$-dimensional vector
\begin{equation}\label{KernelSHAP solution}
\widetilde{\bphi}=
\left(\E_{\bZ \sim p}\left[\left(\widetilde{\bZ}-Z_q\widetilde{\bOne}\right)  \left(\widetilde{\bZ}-Z_q\widetilde{\bOne}\right)^\top \right]\right)^{-1}
\E_{\bZ \sim p}
\left[ \left(\nu(\bZ)-Z_q \nu(\bOne)\right)\left(\widetilde{\bZ}-Z_q\widetilde{\bOne}\right)\right],
\end{equation}
and we set for the $q$-th component $\phi_q = \nu(\bOne)- \widetilde{\bOne}^\top \widetilde{\bphi}$.
These are the exact Shapley values \eqref{Shapely decomposition} for the given value function $\nu$.

\medskip

For later purposes we define the following two matrices
\begin{eqnarray}\label{matrix I}
{\cal I}&=&
\E_{\bZ \sim p}
\left[ \left(\nu(\bZ) -Z_q \nu(\bOne)- \left(\widetilde{\bZ}-Z_q\widetilde{\bOne}\right)^\top \widetilde{\bphi}\right)^2\left(\widetilde{\bZ}-Z_q\widetilde{\bOne}\right)\left(\widetilde{\bZ}-Z_q\widetilde{\bOne}\right)^\top\right],
\\\label{matrix J}
{\cal J}&=&
\E_{\bZ \sim p}
\left[ \left(\widetilde{\bZ}-Z_q\widetilde{\bOne}\right)\left(\widetilde{\bZ}-Z_q\widetilde{\bOne}\right)^\top\right].
\end{eqnarray}
The matrix ${\cal I}$ considers the expected value of the squared gradient computed as in \eqref{KernelSHAP 4} and ${\cal J}$ considers the expected Hessian w.r.t.~$\widetilde{\bphi}$.

\subsection{Sampling KernelSHAP}
To deal with the combinatorial complexity for large $q$, Lundberg--Lee \cite{lundberg2017} propose an empirical version of \eqref{KernelSHAP 4}. Generate an i.i.d.~sample $\bZ_i \sim p$, $1\le i \le n$, and study the empirical Z-estimation problem 
\begin{equation}\label{SHAP 5Z}
\frac{1}{n} \sum_{i=1}^n  \left(\nu(\bZ_i) - \left(\widetilde{\bZ}_i-Z_{i,q}\widetilde{\bOne}\right)^\top \widehat{\bphi}_n-Z_{i,q} \nu(\bOne)\right)\left(\widetilde{\bZ}_i-Z_{i,q}\widetilde{\bOne}\right)~=~\widetilde{\b0}.
\end{equation}
Based on the i.i.d.~sample $(\bZ_i)_{i=1}^n$, define the response vector
and the design matrix, respectively,
\begin{eqnarray}\nonumber
\bY&=&\left(\nu(\bZ_1) -Z_{1,q} \nu(\bOne), \ldots, \nu(\bZ_n) -Z_{n,q} \nu(\bOne)\right)^\top ~\in~ \R^{n},
\\\label{design Z}
{\mathfrak Z}&=&\left(\widetilde{\bZ}_1-Z_{1,q} \widetilde{\bOne}, \ldots, \widetilde{\bZ}_n-Z_{n,q} \widetilde{\bOne}\right)^\top ~\in~ \R^{n\times(q-1)}.
\end{eqnarray}
This allows us to solve \eqref{SHAP 5Z}, which gives the `Sampling KernelSHAP' estimates
\begin{equation}\label{Sample KernelSHAP}
\widehat{\bphi}_n = \left({\mathfrak Z}^\top {\mathfrak Z}\right)^{-1}
{\mathfrak Z}^\top \bY ~\in~\R^{q-1}.
\end{equation}
Note that \eqref{Sample KernelSHAP} requires that the design matrix
${\mathfrak Z}$ has full rank $q-1$ to receive a unique solution $\widehat{\bphi}_n$
for \eqref{SHAP 5Z}.
Using the law of large numbers (LLN), we know that this empirical solution \eqref{Sample KernelSHAP} converges to the true one given in \eqref{KernelSHAP solution}, i.e., we have strict consistency of the estimator $\widehat{\bphi}_n$ for $\widetilde{\bphi}$ for increasing sample size $n\to \infty$; see 
Covert--Lee \cite[Section 4.1]{covert2021}.

\begin{rem}[unbiasedness]\normalfont
Strict consistency of $\widehat{\bphi}_n$ for $\widetilde{\bphi}$ is an asymptotic result, and one may raise the question whether $\widehat{\bphi}_n$ is unbiased for $\widetilde{\bphi}$. This question needs to be denied. Note that for any $n \ge 1$, with positive probability the design matrix ${\mathfrak Z} \in \R^{n \times (q-1)}$ does not have full rank $q-1$. Therefore, with positive probability, $\widehat{\bphi}_n$ is not unique, and hence we cannot speak about unbiasedness because in these cases there is not a well-defined single solution to \eqref{SHAP 5Z}. In this sense, we deny the open question in Covert--Lee \cite[Section 4.1]{covert2021}
and Merrick--Taly \cite{MerrickTaly} of $\widehat{\bphi}_n$ being unbiased.
\EndRemark
\end{rem}
\medskip

Another open question in Covert--Lee \cite[Section 4.1]{covert2021} concerns the asymptotic behavior and the speed of convergence of $\widehat{\bphi}_n$ to $\widetilde{\bphi}$. The following new result gives the answer. 
Assume $\nu(\bZ)$ is not a linear form. 
Using results from Gourieroux et al.~\cite[Appendix 1]{GourierouxMontfortTrogon},
Gallant--Holly \cite{GallantHolly} and Burguete et al.~\cite{Burguete},
one establishes the asymptotic normality result
  \begin{equation}\label{asymptotics of KernelSHAP}
    \sqrt{n} \left(\widehat{\bphi}_n- \widetilde{\bphi}\right)~ \Longrightarrow~ {\cal N}\left(\b0, {\cal T} \right) \qquad \text{ for $n\to \infty$;}
\end{equation}
with asymptotic covariance matrix ${\cal T}={\cal J}^{-1}{\cal I} {\cal J}^{-1}$, and the two matrices ${\cal I}$ and ${\cal J}$ being introduced in 
\eqref{matrix I}-\eqref{matrix J}. This confirms the conjecture in 
Covert--Lee \cite[Section 4.1]{covert2021} of a convergence rate of $O(\sqrt{n})$. Moreover, the matrix ${\cal T}$ specifies the explicit constants of the rate of convergence.

If the value function $\nu$ takes a linear form 
\begin{equation}\label{linear form}
\nu(\bZ)=\bbeta^\top \bZ,
\end{equation}
for a given $\bbeta \in \R^q$, 
we have Shapley value $\bphi=\bbeta$, and its estimate $\widehat{\bphi}_n=\widetilde{\bbeta}$ is exact as soon as ${\mathfrak Z}$ has full rank $q-1$.

\begin{example}[asymptotics of Sampling KernelSHAP]\normalfont
\label{Example 1}
We select a small scale example to verify the asymptotic result \eqref{asymptotics of KernelSHAP}. We choose $q=4$, which results in $2^q-1=15$ non-empty coalitions ${\cal C} \subseteq {\cal Q}=\{1,\ldots, 4\}$. These coalitions are represented by the binary vectors
$\bZ=(Z_1,\ldots, Z_4)^\top \in \{0,1\}^4$. Moreover, we select the (non-linear, normalized) value function
$\nu(\bZ)=\exp ( \bZ^\top \bbeta)-1$ with $\bbeta=(-0.5,  0.1,  0.8, -0.2)^\top$.

In this small scale example the non-zero probability weights $p({\cal C})$ have cardinality $2^4-2=14$, and \eqref{KernelSHAP 3} can easily be computed giving us the (exact) Shapley values
\begin{equation*}
\bphi = ( -0.6025740,  0.1194994,  0.9445458, -0.2400684)^\top.
\end{equation*}
This exact Shapley values are used to benchmark the Sampling KernelSHAP approximations obtained by \eqref{Sample KernelSHAP} for different sample sizes $n \in \N$ and different (selected) realizations of the sample $(\bZ_i)_{i=1}^n$ all having full rank $q-1$; the `selected' refers to the full rank property.

\begin{figure}[htb!] 
\begin{center}
\begin{minipage}{0.32\textwidth}
\begin{center}
\includegraphics[width=\linewidth]{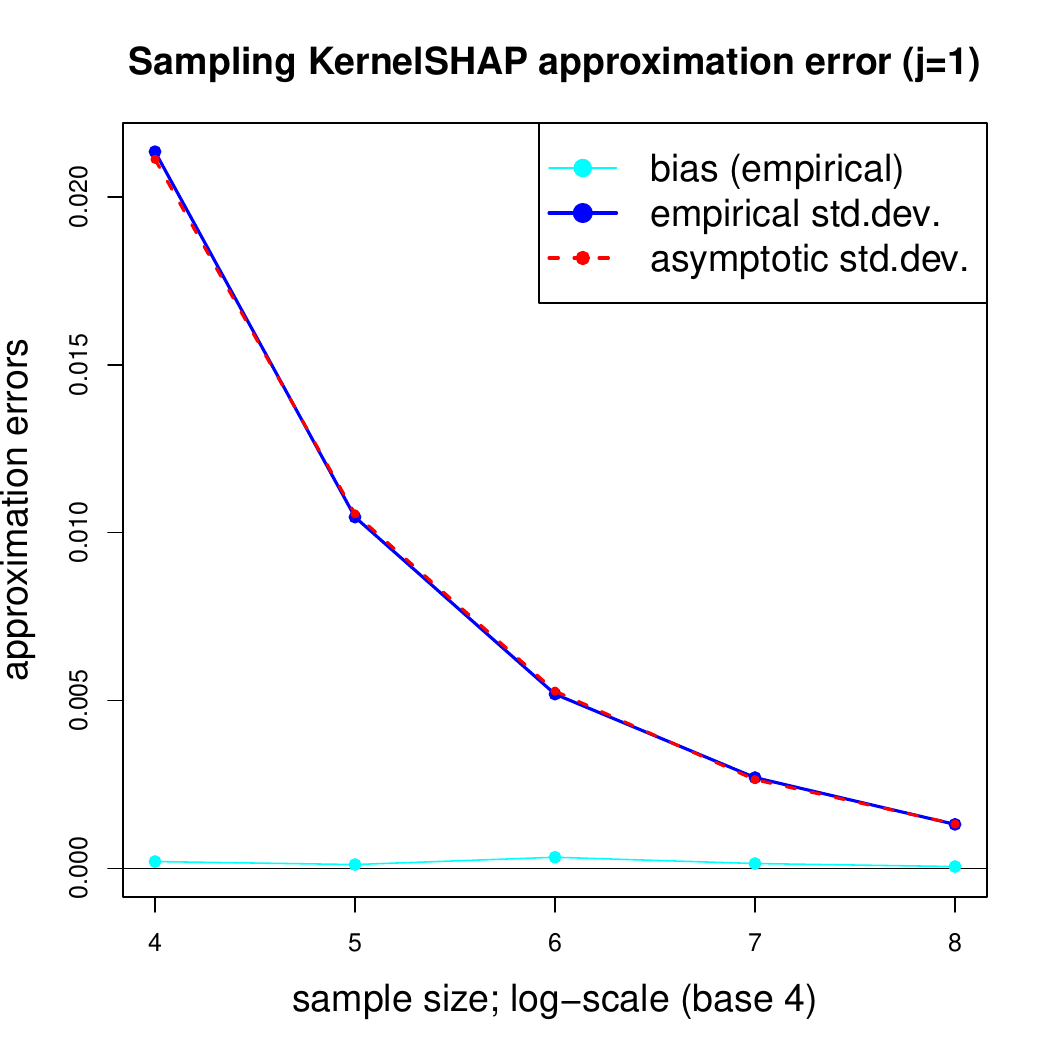}
\end{center}
\end{minipage}
\begin{minipage}{0.32\textwidth}
\begin{center}
\includegraphics[width=\linewidth]{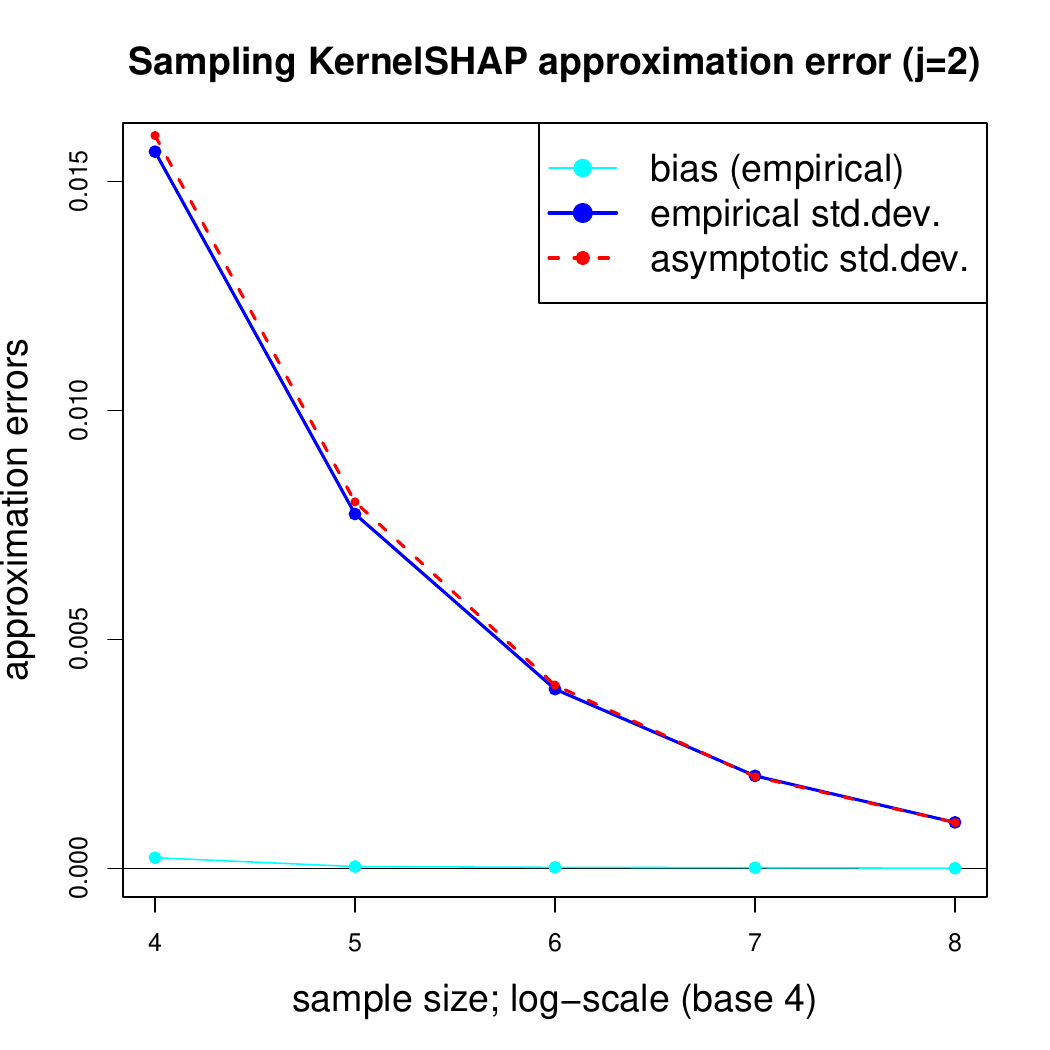}
\end{center}
\end{minipage}
\begin{minipage}{0.32\textwidth}
\begin{center}
\includegraphics[width=\linewidth]{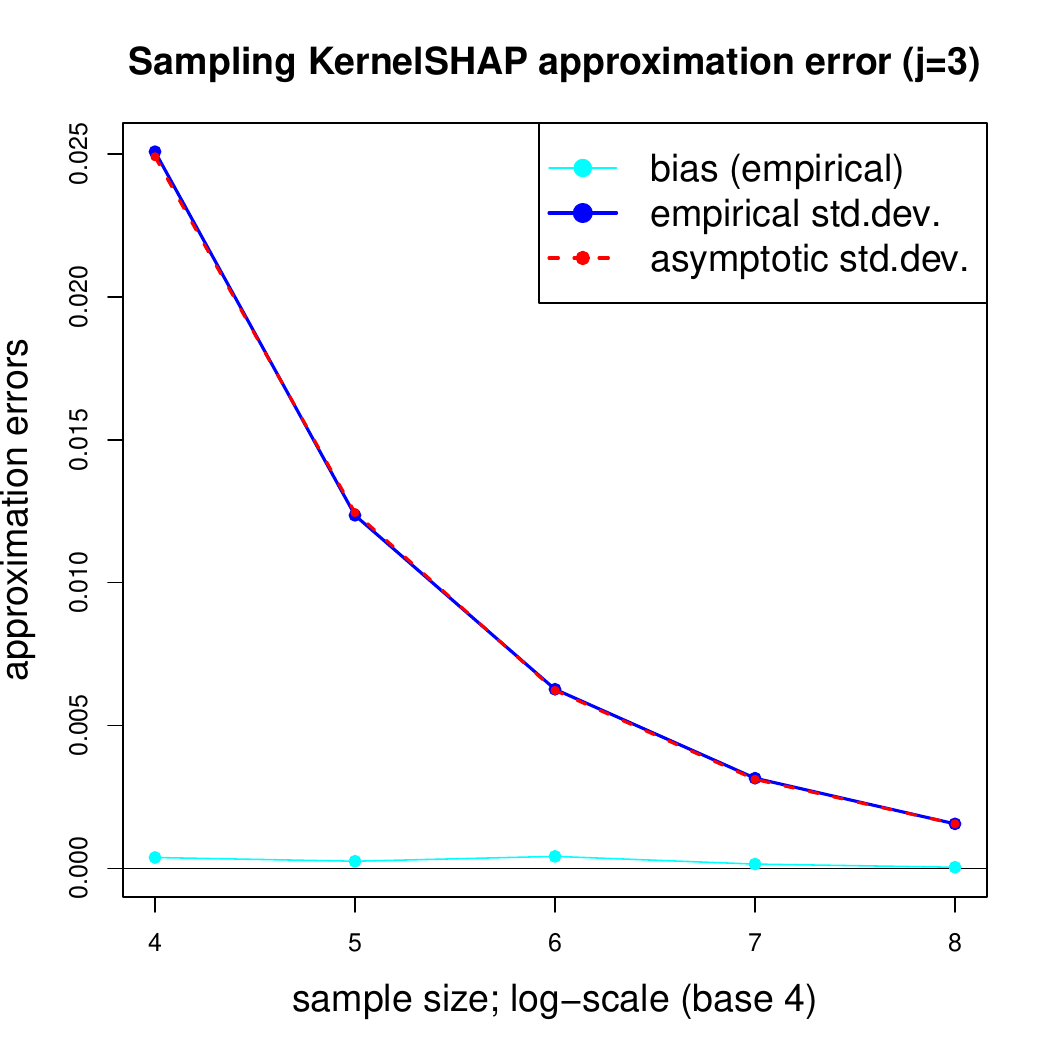}
\end{center}
\end{minipage}
\end{center}
\caption{Sampling KernelSHAP approximation error and uncertainty (standard deviation) as a function of the sample size $n$ for the three components $\widetilde{\bphi}=(\phi_1,\phi_2,\phi_3)^\top$ (lhs-middle-rhs).}
\label{Figure 1}
\end{figure}

Figure \ref{Figure 1} shows the results. We select the sample sizes (on the $x$-axis of 
Figure \ref{Figure 1})
\begin{equation}
n \in {\cal N}=\{4^4, 4^5,4^6,4^7,4^8\}=\{  256,  1'024,  4'096, 16'384, 65'536\},
\end{equation}
to compute the solution $\widehat{\bphi}_n$ of \eqref{SHAP 5Z} by simulating realizations of the sample $(\bZ_i)_{i=1}^n$. We repeat this $S=1000$ times, providing 1000 estimates $\widehat{\bphi}^{(s)}_n$, $1\le s \le S$. The cyan graph shows the resulting average empirical biases
\begin{equation}\label{average empirical bias}
\widehat{\operatorname{bias}}_n(j) = \frac{1}{S} \sum_{s=1}^S \left|
\widehat{\phi}^{(s)}_{n,j} - \phi_j \right|,
\end{equation}
of the components $j \in \{1,\ldots, 2^q-1\}$ for the sample sizes $n\in {\cal N}$. These average empirical biases are negligible in this example. This is often observed as mentioned in Covert--Lee \cite{covert2021}. 

Next, we compute the empirical standard deviations
\begin{equation*}
\widehat{\sigma}_n(j) = \sqrt{\frac{1}{S-1} \sum_{s=1}^S \left(
\widehat{\phi}^{(s)}_{n,j} - \frac{1}{S} \sum_{s=1}^S \widehat{\phi}^{(s)}_{n,j}\right)^2},
\end{equation*}
these empirical standard deviations quantify the average estimation error.
These empirical standard deviations $(\widehat{\sigma}_n(j))_{n\in {\cal N}}$ are plotted in blue color in Figure \ref{Figure 1}, and they show a square root decrease $\sqrt{n}$ as a function of the sample size $n$.

Finally, we use these results to benchmark the standard deviation approximations obtained from the asymptotic normality result \eqref{asymptotics of KernelSHAP}. The red dotted line shows the scaled and square-rooted diagonal of the matrix ${\cal T}={\cal J}^{-1}{\cal I} {\cal J}^{-1} $, that is, 
\begin{equation*}
\tau_n(j) = \sqrt{{\cal T}_{j,j}} / \sqrt{n}.
\end{equation*}
 Comparing $(\widehat{\sigma}_n(j))_{n\in {\cal N}}$ and $(\tau_n(j))_{n\in {\cal N}}$ we observe a good alignment which supports an uncertainty estimation by the 
asymptotic normality result \eqref{asymptotics of KernelSHAP}.
\EndExample
\end{example}

We conclude that in our example, the asymptotic normality result gives a very accurate estimate of the approximation error of the Sampling KernelSHAP, already for the small sample size of
$n=4^4=256$. This very useful information because it tells us that the asymptotic covariance matrix is capable to give rather precise error estimates already for small sample sizes.
To apply the asymptotic normality result in practice, one needs estimates for the matrices ${\cal I}$ and ${\cal J}$. This can be done empirically by using a realization of $(\bZ_i)_{i=1}^n$, and for matrix ${\cal I}$ we additionally use the estimate $\widehat{\bphi}_n$.

\subsection{Paired-Sampling KernelSHAP}
The reduce the variance and improve the approximation, 
Covert--Lee \cite[Section 4.2]{covert2021} propose to perform paired-sampling.
This means that for every instance $\bZ \sim p$ one simultaneously considers its 
complement $\bZ^c= \bOne-\bZ$ which has the same probability weight as $\bZ$. This allows one to modify \eqref{KernelSHAP 2} to the paired minimization
\begin{equation*}\underset{\bphi \in \R^q}{\arg\min}\,\E_{\bZ \sim p}
\left[ \left(\nu(\bZ) - \bZ^\top \bphi\right)^2+\left(\nu(\bZ^c) - (\bZ^c)^\top \bphi\right)^2\right]
\qquad \text{ subject to $\bOne^\top\bphi = \nu(\bOne)$.}
\end{equation*}
Inserting the side constraint and computing the Z-estimation problem 
gives the score equations
\begin{equation}\label{scores 2}
\E_{\bZ \sim p}
\left[2 \left(\frac{\nu(\bZ)+\nu(\bOne)-\nu(\bOne-\bZ)}{2}-Z_q \nu(\bOne) - \left(\widetilde{\bZ}-Z_q\widetilde{\bOne}\right)^\top \widetilde{\bphi}\right)\left(\widetilde{\bZ}-Z_q\widetilde{\bOne}\right)\right]=\widetilde{\b0}.
\end{equation}
This can be solved in complete analogy to \eqref{KernelSHAP solution}, and it gives the exact Shapley values.

\medskip

We define the following two matrices
 {\small
\begin{eqnarray*}
{\cal I}_2&=&
\E_{\bZ \sim p}
\left[4\left(\frac{\nu(\bZ)+\nu(\bOne)-\nu(\bOne-\bZ)}{2}-Z_q \nu(\bOne) - \left(\widetilde{\bZ}-Z_q\widetilde{\bOne}\right)^\top \widetilde{\bphi}\right)^2\left(\widetilde{\bZ}-Z_q\widetilde{\bOne}\right)\left(\widetilde{\bZ}-Z_q\widetilde{\bOne}\right)^\top\right],
\\
{\cal J}_2&=&2\,
\E_{\bZ \sim p}
\left[ \left(\widetilde{\bZ}-Z_q\widetilde{\bOne}\right)\left(\widetilde{\bZ}-Z_q\widetilde{\bOne}\right)^\top\right]~=~2\,{\cal J}.
\end{eqnarray*}
}

The `Paired-Sampling KernelSHAP' estimate $\widehat{\bphi}_n^{(PS)}$ is found completely analogously to \eqref{SHAP 5Z} by simultaneously considering 
the i.i.d.~instances $\bZ_i$ and their complements $\bZ_i^c=\bOne -\bZ_i$ in the design matrix.
Based on Gourieroux et al.~\cite[Appendix 1]{GourierouxMontfortTrogon},
Gallant--Holly \cite{GallantHolly} and Burguete et al.~\cite{Burguete}
one establishes the following result of asymptotic normality
for the paired-sampling estimate
\begin{equation}\label{paired asymptotics of KernelSHAP}
    \sqrt{n} \left(\widehat{\bphi}^{(PS)}_n- \widetilde{\bphi}\right)~ \Longrightarrow~ {\cal N}\left(\b0, {\cal T}_2 \right) \qquad \text{ for $n\to \infty$;}
\end{equation}
for asymptotic covariance matrix ${\cal T}_2={\cal J}_2^{-1}{\cal I}_2 {\cal J}_2^{-1}$ and
supposed $\nu(\bZ)$ is not a bilinear form. This last clause will be further explained below.

\medskip

The following proposition proves that paired-sampling is beneficial.
\begin{prop}\label{Prop2.3}
We have positive semi-definite matrix
\begin{equation*}
{\cal T}-{\cal T}_2=
{\cal J}^{-1}{\cal I}{\cal J}^{-1}-{\cal J}_2^{-1}{\cal I}_2 {\cal J}_2^{-1}~\ge~ 0.
\end{equation*}
\end{prop}
This result is proved in the appendix.

\begin{example}[Example \ref{Example 1}, revisited]
\label{Example 2}\normalfont
We compare the Paired-Sampling KernelSHAP to its non-paired counterpart in the same set-up as in Example \ref{Example 1}.

\begin{figure}[htb!] 
\begin{center}
\begin{minipage}{0.32\textwidth}
\begin{center}
\includegraphics[width=\linewidth]{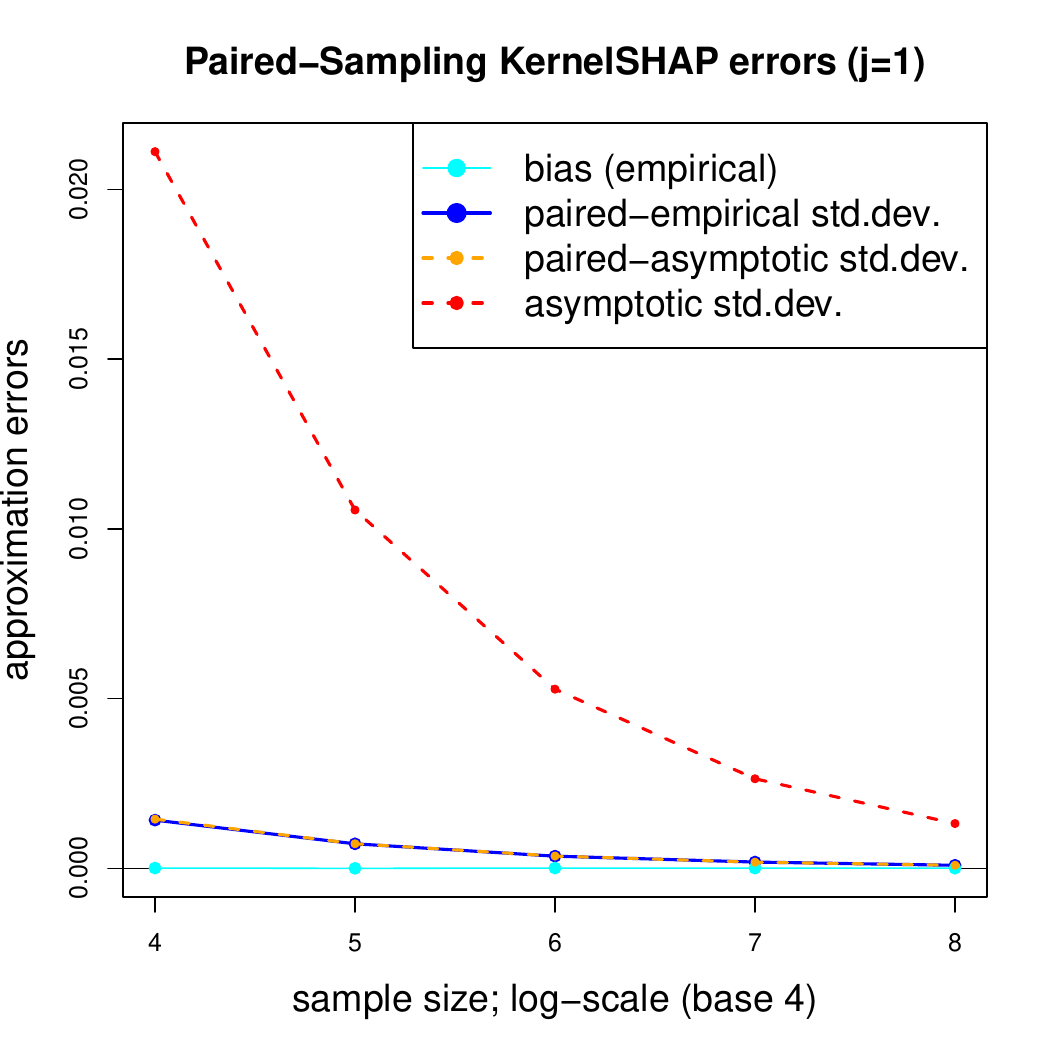}
\end{center}
\end{minipage}
\begin{minipage}{0.32\textwidth}
\begin{center}
\includegraphics[width=\linewidth]{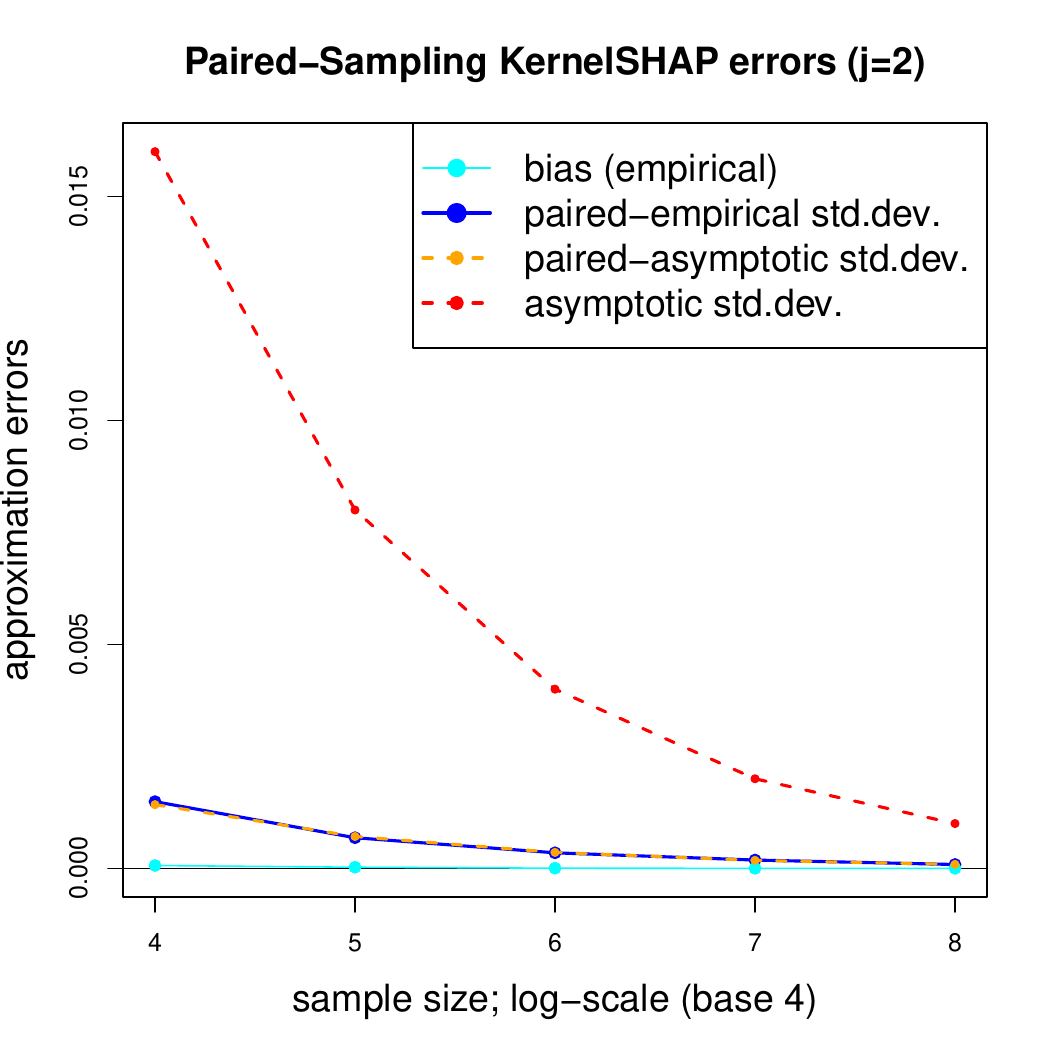}
\end{center}
\end{minipage}
\begin{minipage}{0.32\textwidth}
\begin{center}
\includegraphics[width=\linewidth]{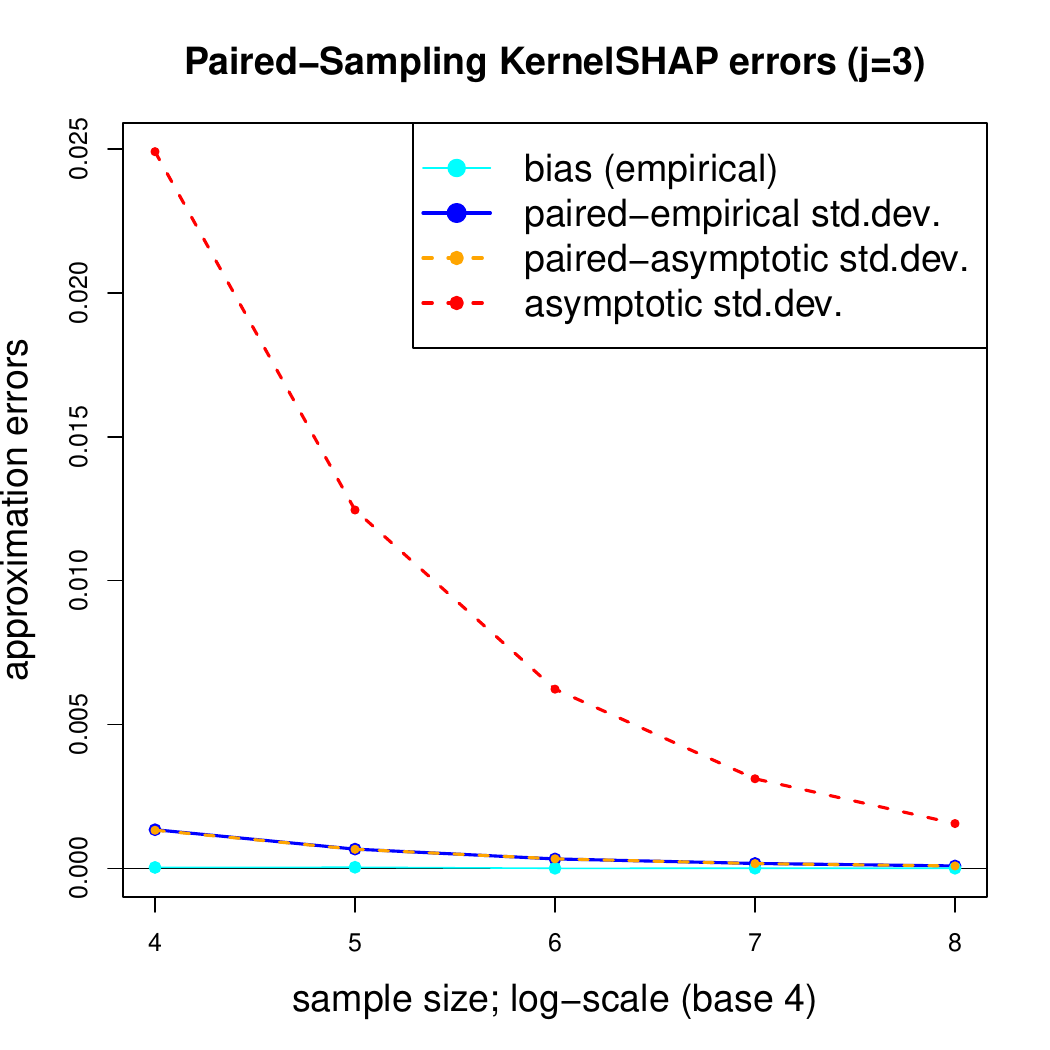}
\end{center}
\end{minipage}
\end{center}
\caption{Paired-Sampling KernelSHAP approximation $\widehat{\bphi}^{(PS)}_n$ and its uncertainty (standard deviation) as a function of the sample size $n$ for the three components $\widetilde{\bphi}=(\phi_1,\phi_2,\phi_3)^\top$ (lhs-middle-rhs); the true Shapley values are 
$\widetilde{\bphi} = ( -0.6025740,  0.1194994,  0.9445458)^\top$.}
\label{Figure 2}
\end{figure}

Figure \ref{Figure 2} verifies that the asymptotic approximation 
\eqref{paired asymptotics of KernelSHAP} is very accurate in our example (compare the orange dotted line to its empirical counterpart in blue). Moreover, it verifies
Proposition \ref{Prop2.3} by showing a significant improvement of the paired version (orange dotted line) over the non-paired one (red dotted line); already with sample size $n=4^4=256$ we receive very accurate estimates on average in this example.
\EndExample
\end{example}

\subsection{Interactions of order 2}
The asymptotic result \eqref{paired asymptotics of KernelSHAP}  holds if the value function $\nu(\bZ)$ is not a bilinear form. The Paired-Sampling KernelSHAP is exact in case of a bilinear form as soon as the design matrix
${\mathfrak Z}$ has full rank $q-1$. We prove this result.

Assume the value function is a bilinear form
\begin{equation}\label{binary interactions}
\nu(\bZ)= \sum_{1\le j , k \le q} Z_j Z_k a_{j,k}
= \bZ^\top A \bZ,
\end{equation}
with matrix $A=\left(a_{j,k}\right)_{1\le j, k \le q} \in \R^{q \times q}$. This means that the value function only involves terms (interactions)  of maximal order 2.

We come back to the score equations in the paired-sampling case \eqref{scores 2}, and we insert the bilinear form \eqref{binary interactions} of the value function.
This gives us the identity
\begin{equation*}
\nu(\bZ)+\nu(\bOne)-\nu(\bOne-\bZ)
=\bZ^\top A \bOne + \bOne^\top A \bZ= 2 \bbeta^\top \bZ,
\end{equation*}
with vector $\bbeta = (A+A^\top) \bOne /2 \in \R^q$. Inserting this into \eqref{scores 2}
gives us the score equations in the bilinear case
\begin{equation*}
\E_{\bZ \sim p}
\left[2\left(\bbeta^\top \bZ -Z_q \bOne^\top \bbeta - \left(\widetilde{\bZ}-Z_q\widetilde{\bOne}\right)^\top \widetilde{\bphi}\right)\left(\widetilde{\bZ}-Z_q\widetilde{\bOne}\right)\right]=\widetilde{\b0}.
\end{equation*}
This aligns with the linear form \eqref{linear form}, and we have the following immediate corollary. 

\begin{cor} \label{cor bilinear}
Assume the value function $\nu$ takes a bilinear form 
\eqref{binary interactions}. The Shapley values are given by
\begin{equation}\label{bilinear Shapleys}
\phi_j = \frac{1}{2}  \sum_{k=1}^q \left(a_{j,k} + a_{k,j}\right)
\qquad \text{ for $j\in {\cal Q}$.}
\end{equation}
\end{cor}

This result also sheds an interesting light on the Paired-Sampling KernelSHAP
in the case of a bilinear form \eqref{binary interactions} of the value function.
In this case, we do not need to sample coalitions $\bZ \sim p$, but we can simply take (any) $q-1$ linearly independent coalitions $\bZ_1,\ldots, \bZ_{q-1}$ and their complements $\bZ^c_1,\ldots, \bZ^c_{q-1}$. Solving the Paired-Sampling KernelSHAP \eqref{Sample KernelSHAP} for these $2(q-1)$ coalitions gives the exact Shapley values $\bphi$ given in \eqref{bilinear Shapleys}.
This result also allows one to test for value functions of interactions of maximal order 2. Namely, if for all possible $q-1$ linearly independent coalitions $\bZ_1,\ldots, \bZ_{q-1}$ the Paired-Sampling KernelSHAP gives an identical result, we have a bilinear form, and otherwise not. 

To the best of our knowledge, the result of Corollary \ref{cor bilinear} is new, though such an insight is not completely new. Lundberg \cite{Lundberg2018B} mentions such a property for the Paired-Sampling PermutationSHAP, which we are going to discuss in the next section.

\section{Sampling PermutationSHAP}
\label{sec Sampling PermutationSHAP}

We analyze the same questions in case of the PermutationSHAP representation
\eqref{permutation SHAP} of \v{S}trumbelj--Kononenko \cite{Strumbelj2010, Strumbelj2014}. The sampling version of the PermutationSHAP is obtained by sampling i.i.d.~permutations $(\pi^{(i)})_{i=1}^n$ of the ordered set $(1,\ldots, q)$.
Again taking advantage of paired-sampling, we simultaneously consider the reverted permutations $\rho(\pi^{(i)})=(\pi^{(i)}_q,\ldots, \pi^{(i)}_1)$. This motivates the
`Paired-Sampling PermutationSHAP' estimate for $j \in {\cal Q}$
\begin{equation}\label{def PS Perm}
\widehat{\phi}_j^{(n)} = \frac{1}{2n} \sum_{i=1}^n \left(\nu\left({\cal C}_{\pi^{(i)}, j}\cup \{j\}\right)
-\nu\left({\cal C}_{\pi^{(i)}, j}\right)
+\nu\left({\cal C}_{\rho(\pi^{(i)}), j}\cup \{j\}\right)
-\nu\left({\cal C}_{\rho(\pi^{(i)}), j}\right) \right).
\end{equation}

In Lundberg \cite{Lundberg2018B}, it is stated that the Paired-Sampling PermutationSHAP is exact in the case of a value function with maximal order 2. We could not find any proof for this claim in the literature, therefore, we state and prove it in the following proposition; the proof is provided in the appendix.

\begin{prop} \label{bilinear SHAP}
In the case of a bilinear form \eqref{binary interactions} for the value function $\nu$ we have 
\begin{equation*}
\widehat{\phi}^{(n)}_j=\phi_j = \frac{1}{2}  \sum_{k=1}^q \left(a_{j,k} + a_{k,j}\right),
\end{equation*}
for $n=1$ and any permutation $\pi=\pi^{(1)}$.
\end{prop}

Corollary \ref{cor bilinear} and Proposition \ref{bilinear SHAP} give independent results for identifying bilinear forms, i.e., value functions with interactions of maximal order 2. The latter result is even simpler because it only needs one single permutation, in particular, $\pi=(1,\ldots, q)$ and its reverted version do the job, i.e., in case of a bilinear value function they provide the exact Shapley values.

\medskip

Finally, we give some asymptotic statements for the Paired-Sampling PermutationSHAP. We rewrite the PermutationSHAP as
\begin{equation*}
\bphi = \E_{\pi} \left[ \bB_\pi \right]
= \frac{1}{2}\,\E_{\pi} \left[ \bB_\pi+\bB_{\rho(\pi)} \right]
\quad \text{ for random vector }
\bB_\pi=
\begin{pmatrix}
\nu\left({\cal C}_{\pi, 1}\cup \{1\}\right)-\nu\left({\cal C}_{\pi, 1}\right)
\\\vdots
\\
\nu\left({\cal C}_{\pi, q}\cup \{q\}\right)-\nu\left({\cal C}_{\pi, q}\right)
\end{pmatrix},
\end{equation*}
where the expectation operator $\E_\pi$ assigns probability $1/q!$ to all permutations $\pi$. Moreover, define the covariance matrix of $\bB_\pi+\bB_{\rho(\pi)}$
by
\begin{equation}\label{permutation Sigma}
\Sigma=\frac{1}{4}\,  {\rm Var}_\pi\left(\bB_\pi+\bB_{\rho(\pi)}\right) ~\in ~\R^{q\times q}.
\end{equation}

There are the following straightforward statements; see also Castro et al.~\cite{Castro}, Maleki et al.~\cite{Maleki} and \v{S}trumbelj--Kononenko \cite{Strumbelj2014}.
The Paired-Sampling PermutationSHAP estimate $\widehat{\phi}^{(n)}_j$
is strictly consistent for $\phi_j$ as $n\to \infty$, it is unbiased for $\phi_j$, and it satisfies the central limit theorem (CLT), i.e., asymptotic normality
\begin{equation}\label{CLT Permutation}
\sqrt{n}
\left((\widehat{\phi}^{(n)}_1, \ldots, \widehat{\phi}^{(n)}_q)^\top-
\bphi\right)~ \Longrightarrow~ {\cal N}\left(\b0, \Sigma \right) \qquad \text{ for $n\to \infty$.}
\end{equation}

\begin{example}[Example \ref{Example 2}, revisited]
\label{Example 3}\normalfont
In this example we compare the performance of the Paired-Sampling KernelSHAP method and Paired-Sampling PermutationSHAP method for the same value function as in Examples \ref{Example 1} and \ref{Example 2}.

\begin{figure}[htb!] 
\begin{center}
\begin{minipage}{0.32\textwidth}
\begin{center}
\includegraphics[width=\linewidth]{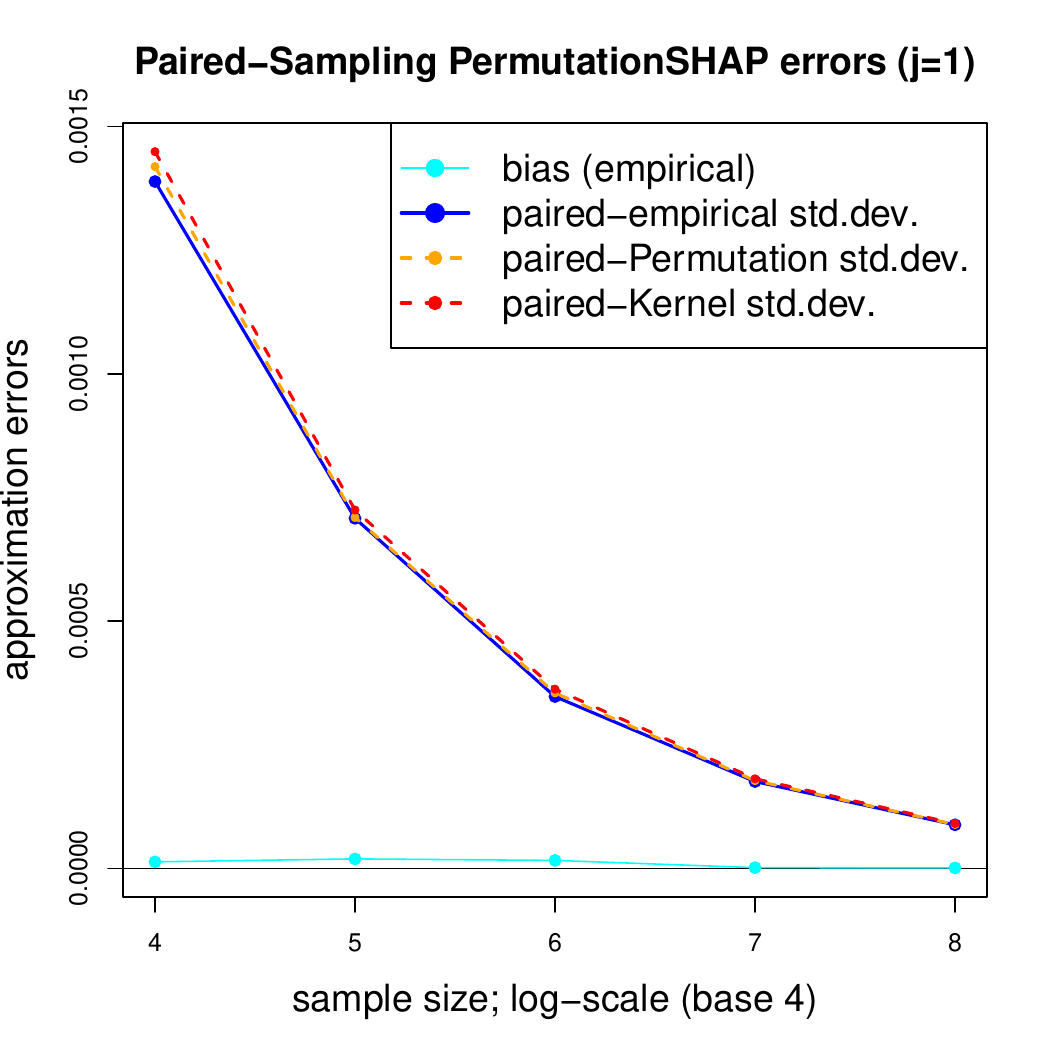}
\end{center}
\end{minipage}
\begin{minipage}{0.32\textwidth}
\begin{center}
\includegraphics[width=\linewidth]{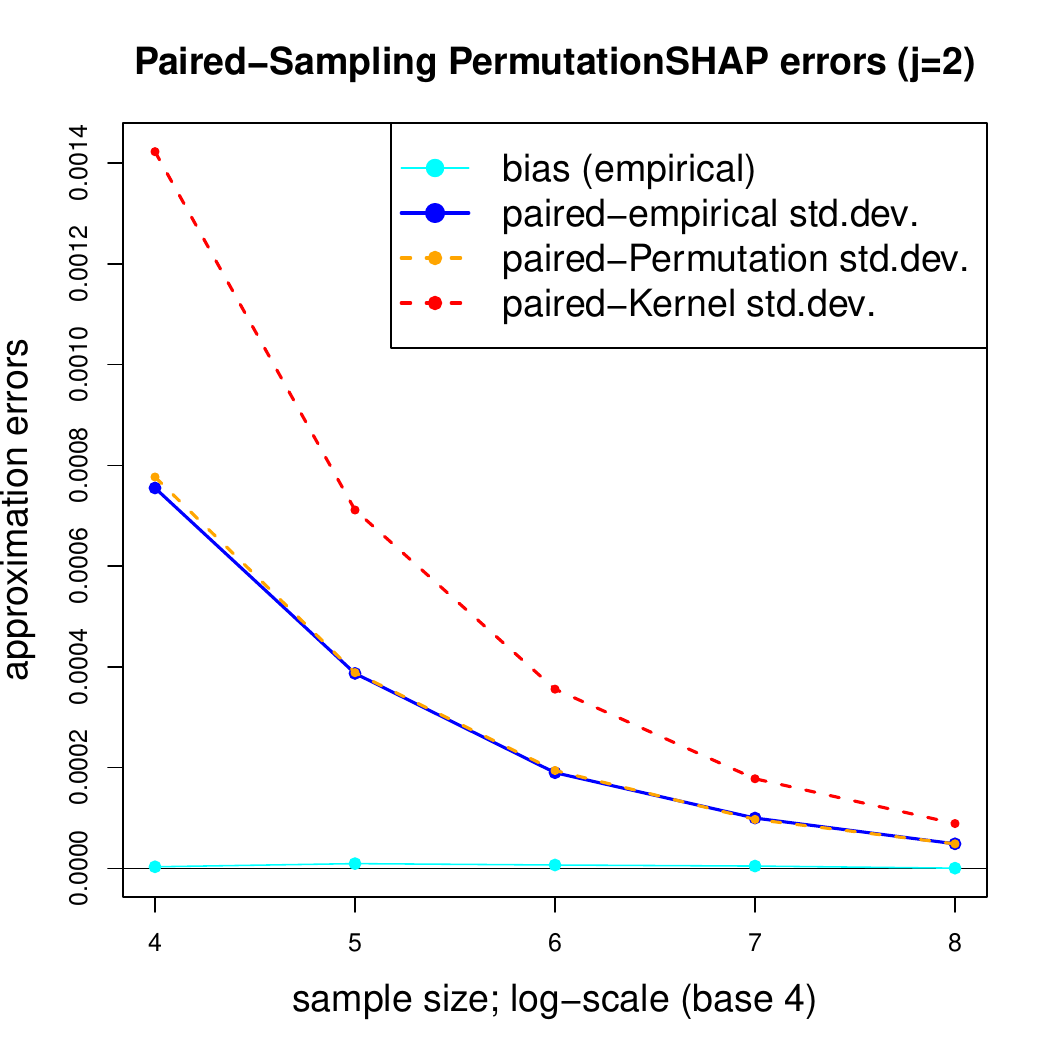}
\end{center}
\end{minipage}
\begin{minipage}{0.32\textwidth}
\begin{center}
\includegraphics[width=\linewidth]{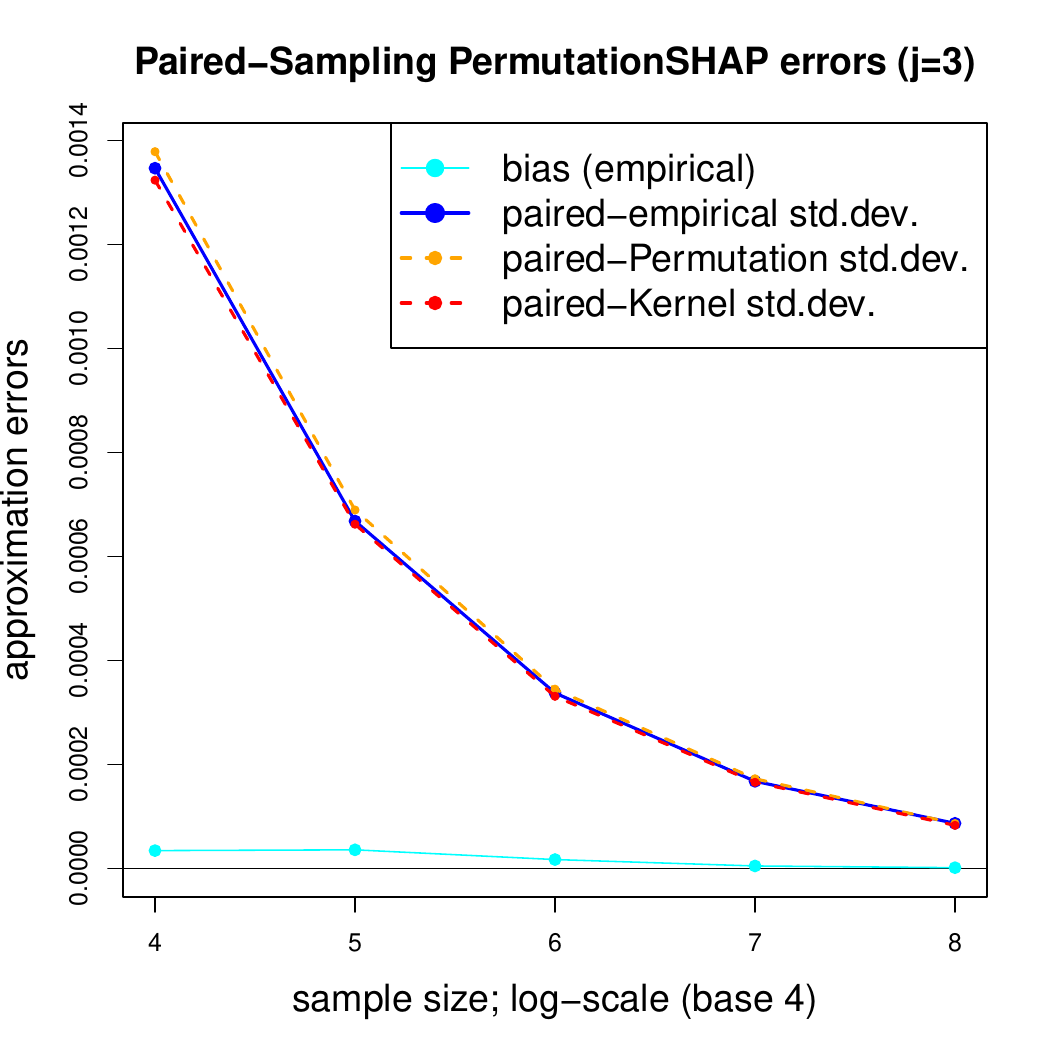}
\end{center}
\end{minipage}
\end{center}
\caption{Paired-Sampling PermutationSHAP approximation $\widehat{\bphi}^{(n)}$ and its uncertainty (standard deviation) as a function of the sample size $n$ for the first three components (lhs-middle-rhs).}
\label{Figure 3}
\end{figure}

The results are shown in Figure \ref{Figure 3} for the first three components.
Again the asymptotic CLT approximation \eqref{CLT Permutation} (orange dotted line) is very close to the empirical version (blue color) which confirms the appropriateness of the asymptotic approximation. The Paired-Sampling PermutationSHAP (orange dotted line) is also compared to the
Paired-Sampling KernelSHAP (red dotted line). The resulting approximation errors are of a similar magnitude in this example, only for the second Shapley value $\phi_2$ we notice some differences, giving preference to the permutation version. Comparing the positive eigenvalues of the two asymptotic covariance matrices ${\cal T}_2={\cal J}_2^{-1}{\cal I}_2{\cal J}_2^{-1}$ and $\Sigma$, respectively, we receive:
\begin{eqnarray*}
\text{Paired-Sampling KernelSHAP}:&&0.00096, 0.00039, 0.00016;
\\
\text{Paired-Sampling PermutationSHAP}:&&0.00075, 0.00070, 0.00020.
\end{eqnarray*}
At the same time the former gives a slightly smaller trace of 0.00151 vs.~0.00165 of the latter. Thus, altogether the performance of the two methods is roughly equally good in our example. This closes this example.
\EndExample
\end{example}

In practice, there is a preference for the permutation version over the kernel one, but from the previous example, it is unclear whether such a preference can be supported. For this reason, we present a slightly bigger example with $q=10$. The exact Shapley values of this bigger example can still be fully computed on an ordinary computer.

\begin{example}\normalfont
We consider the same set-up as in Examples \ref{Example 1}, \ref{Example 2} and \ref{Example 3}, but we choose a bigger grand coalition $q=10$. For the value function, we select
$\nu(\bZ)=\exp\{ \bZ^\top \bbeta\}$ with the components of $\bbeta \in \R^q$ being selected by i.i.d.~standard Gaussian random variables; this parameter $\bbeta$ is sampled once and kept fixed throughout the entire example. We then perform precisely the same computations for the Paired-Sampling KernelSHAP as in Example \ref{Example 2} and for the Paired-Sampling PermutationSHAP as in Example \ref{Example 3}. 

\begin{figure}[htb!] 
\begin{center}
\begin{minipage}{0.32\textwidth}
\begin{center}
\includegraphics[width=\linewidth]{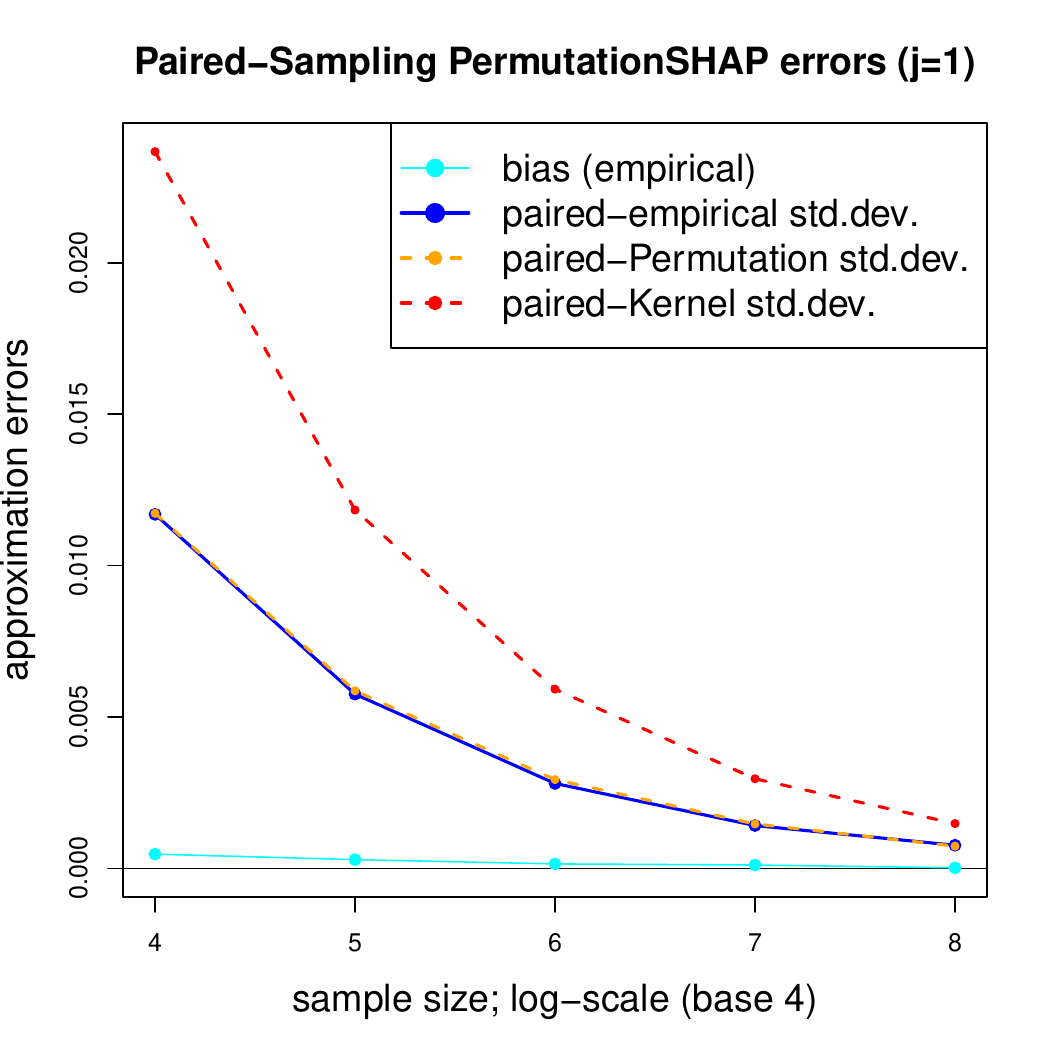}
\end{center}
\end{minipage}
\begin{minipage}{0.32\textwidth}
\begin{center}
\includegraphics[width=\linewidth]{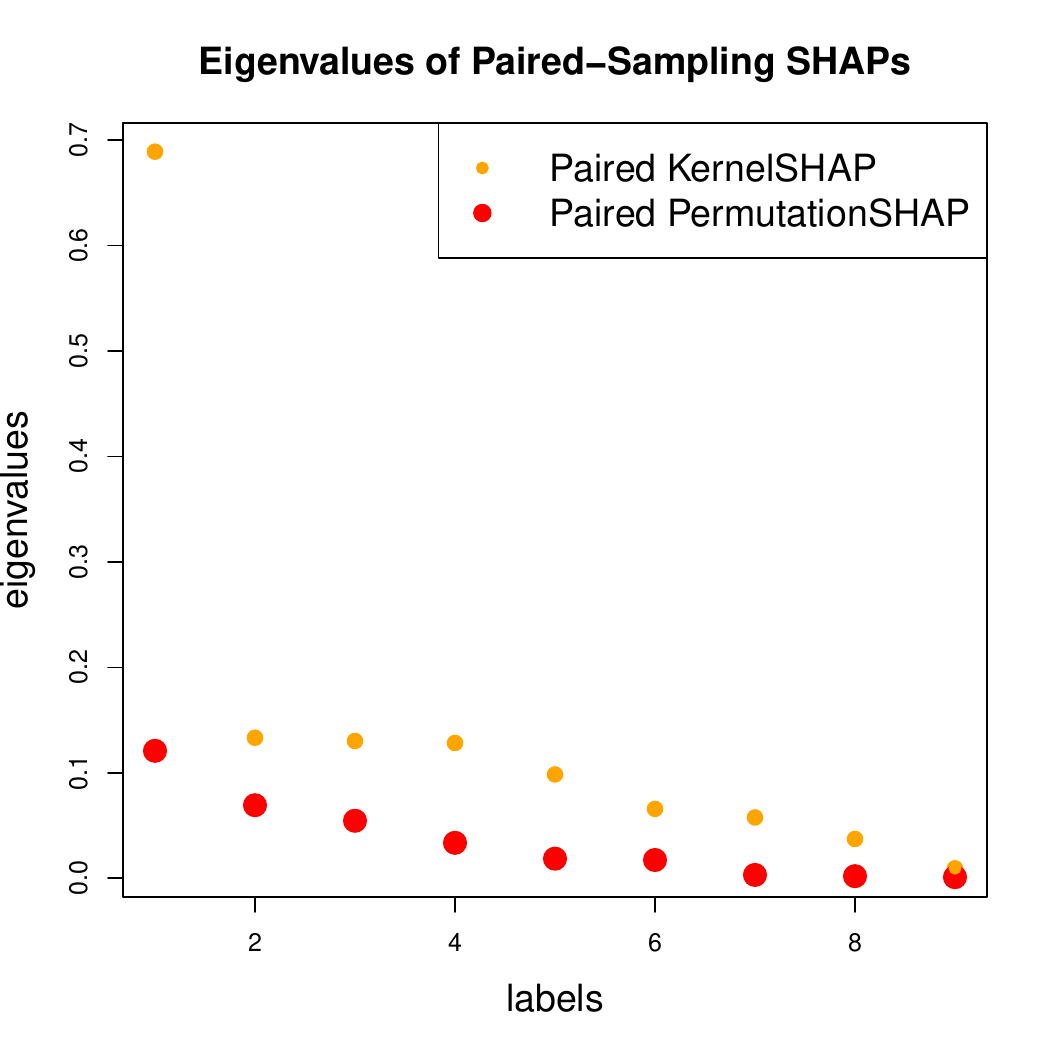}
\end{center}
\end{minipage}
\begin{minipage}{0.32\textwidth}
\begin{center}
\includegraphics[width=\linewidth]{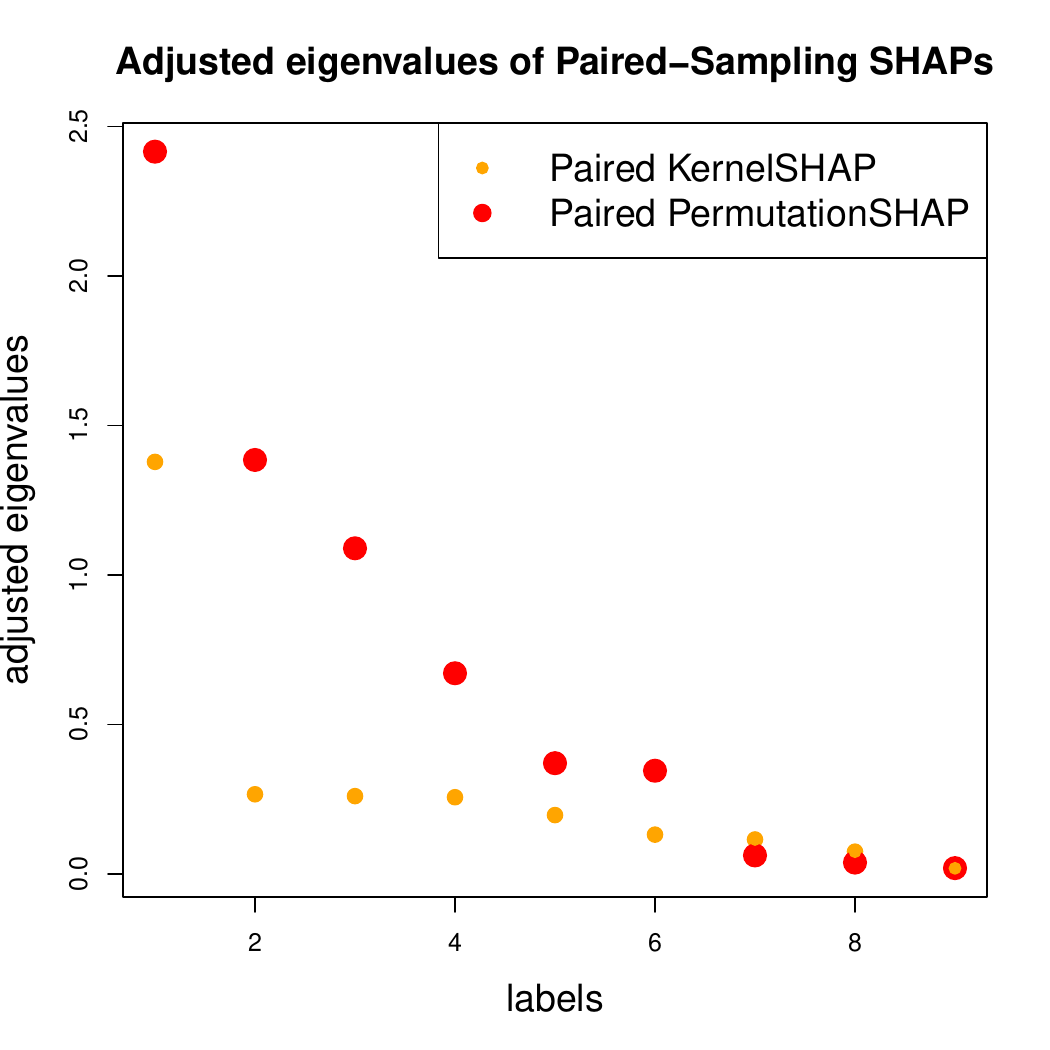}
\end{center}
\end{minipage}
\end{center}
\caption{Paired-Sampling KernelSHAP and PermutationSHAP uncertainties (standard deviations) as a function of the sample size $n$ for the first component $j=1$ (lhs), the eigenvalues of the asymptotic covariance matrices ${\cal T}_2$ and $\Sigma$ (middle), and dimension-adjusted eigenvalues (rhs).}
\label{Figure 4}
\end{figure}

Similarly to this latter example, this allows us to compare the
$q=10$ Shapley value estimates; for illustrative reasons, we only show the
plot of the first one, $j=1$, in Figure \ref{Figure 4} (lhs).
The cyan line shows the empirical bias of the Paired-Sampling PermutationSHAP, this bias is negligible.
The blue line verifies that the asymptotic approximations of the standard errors are very accurate in the Paired-Sampling PermutationSHAP. Finally, the red dotted line shows that the corresponding standard deviations of the Paired-Sampling KernelSHAP are significantly larger, giving a clear preference to the permutation version over the kernel one. Figure \ref{Figure 4} (middle) shows the eigenvalues of the asymptotic covariance matrices ${\cal T}_2$ (Paired-Sampling KernelSHAP) and $\Sigma$ (Paired-Sampling PermutationSHAP). This figure again gives a clear preference to the permutation version.

One may argue that the graph in Figure \ref{Figure 4} (middle) does not study the correct quantity, because it analyzes the asymptotic behavior as a function of the sample size $n$. However, it would be more interesting to compare the computational time to determine these quantities. By far the most expensive operation in complex machine learning models is the evaluation of the value function $\nu$.
In each paired-sample of the KernelSHAP version, we need to evaluate this function twice.
In the case of the Paired-Sampling PermutationSHAP, each simulation requires $2q$ evaluations. If we multiply the eigenvalues with these factors ($2$ and $2q$, respectively), we receive the dimension-adjusted eigenvalues shown in Figure \ref{Figure 4} (rhs). This graph says that if the main trigger is the computational efficiency of the value function evaluation, we should give preference to the Paired-Sampling KernelSHAP version.
\EndExample
\end{example}

\section{Additive recovery property}
\label{Additive recovery property}
From the previous sections, we conclude that there is no clear preference for one of the two paired-sampling SHAP versions. Moreover, both provide the exact solutions for bilinear forms, see Corollary \ref{cor bilinear} and Proposition \ref{bilinear SHAP}. This section extends the results of these bilinear forms to the case where the value function decomposes into additive parts, and we try to recover these additive parts; this is related to the additive recovery property of Apley--Zhu \cite{apley2016}. For these generalized versions of the bilinear form, we will prove that the Paired-Sampling PermutationSHAP possesses this additive recovery property, whereas the Paired-Sampling KernelSHAP does not.

Assume that we can partition the value function $\nu$ as follows 
$$\bZ \mapsto \nu(\bZ) =\nu_1(\bZ) + \nu_2(\bZ),$$
with a bilinear form $\nu_1(\bZ)$ and a general value function $\nu_2(\bZ)$ such that the components of $\bZ$ that influence either one of the two value functions are disjoint. Since we can reorder the components of $\bZ$, we can assume w.l.o.g.~that the first $d$ components of $\bZ$ are involved into the bilinear form, and the remaining $q-d+1$ components are not. We set
\begin{equation*}
\bZ ~=~
\begin{pmatrix}
Z_1\\ \vdots \\ Z_d \\ Z_{d+1}\\ \vdots  \\ Z_q 
\end{pmatrix}
~=~
\begin{pmatrix}
Z_1\\ \vdots \\ Z_d \\ 0\\ \vdots  \\0 
\end{pmatrix}+
\begin{pmatrix}
0\\ \vdots \\ 0 \\ Z_{d+1}\\ \vdots  \\ Z_q 
\end{pmatrix} ~=:~ \bZ_1 + \bZ_2.
\end{equation*}
We then assume that the value function $\nu$ allows for a partition such that for all binary vectors $\bZ$ we can write
\begin{equation}\label{separation value function}
\nu(\bZ) =\nu_1(\bZ) + \nu_2(\bZ)=\nu_1(\bZ_1) + \nu_2(\bZ_2), 
\end{equation}
with the first value function being a bilinear form satisfying
\begin{equation}\label{bilinear form 1}
\nu_1(\bZ) = \bZ^\top A \bZ= \bZ_1^\top A \bZ_1,
\end{equation}
for a matrix $A \in \R^{q \times q}$, where only the upper-left $(d\times d)$-square is different from zero. Similarly, $\nu_2(\bZ)=\nu_2(\bZ_2)$ only involves the $q-d+1$ components, thus, there is no interaction between the first $d$ components and the last $q-d+1$ components of $\bZ$.

\begin{cor} 
\label{cor xz}
Assume that the value function $\nu=\nu_1+\nu_2$ decomposes as in \eqref{separation value function} into a bilinear form  \eqref{bilinear form 1} and a general term $\nu_2$, such that the components that influence either one of the terms are disjoint. We have for the first $d$ Shapley values
\begin{equation*}
\phi_j = \frac{1}{2}  \sum_{k=1}^d \left(a_{j,k} + a_{k,j}\right), 
\qquad \text{ for $1\le j \le d$.}
\end{equation*}
\end{cor}

{\Beweis
{\bf Proof.} The fact that the last $q-d+1$ components of $\bZ$ do not influence the value function $\nu_1$ makes them dummy players for the first value function. Similarly, the fact that the first $d$ components of $\bZ$ do not influence the value function $\nu_2$ makes them dummy players for the second value function.
The claim then immediately follows from the linearity axiom (A4), the dummy axiom (A3) and Corollary \ref{cor bilinear}.
\EndProof
}

\medskip

This corollary and its proof tell us that if the value function $\nu$ additively decomposes into two separate parts \eqref{separation value function} leading to a partition of the components of $\bZ$, then we can solve two independent Shapley decompositions (this follows from axioms (A4) and (A3)), and since the first one for $\nu_1$ is a bilinear form, Corollary \ref{cor bilinear} applies to the first part.
We have the following proposition for the Paired-Sampling PermutationSHAP.

\begin{prop} \label{bilinear SHAP 2}
Assume that the value function $\nu=\nu_1+\nu_2$ decomposes as in \eqref{separation value function} into a bilinear form  \eqref{bilinear form 1} and a general term $\nu_2$, such that the components that influence either one of the terms are disjoint.
We have for the first $d$ Shapley values, $1\le j \le d$, using the Paired-Sampling PermutationSHAP
\begin{equation*}
\widehat{\phi}^{(1)}_j=\phi_j = \frac{1}{2}  \sum_{k=1}^d \left(a_{j,k} + a_{k,j}\right),
\end{equation*}
for any permutation $\pi=\pi^{(1)}$, and where $\widehat{\phi}^{(1)}_j$
is given by \eqref{def PS Perm} with one single permutation $n=1$.
\end{prop}

This proposition tells us that if there exists a subset of components of $\bZ$ that only has interactions of maximal degree two, and no interactions with the other components which may have higher order interactions, then one single paired permutation allows one to exactly compute the Shapley values of these components using the paired-sampling PermutationSHAP. Note that for the second (general) part $\nu_2$ of the value function  we can only make the asymptotic normality statement \eqref{CLT Permutation}.

A similar result does not hold for the Paired-Sampling KernelSHAP, also not for purely additive components without interactions. This can easily be verified by a numerical counterexample. The next example proves that 
Proposition \ref{bilinear SHAP 2} does not hold for the Paired-Sampling KernelSHAP, and it is verified that it holds for the Paired-Sampling PermutationSHAP.

\begin{example}\normalfont
We select the following value function. Select $q=5$ and  choose
\begin{equation*}
\nu(\bZ) = (Z_1,Z_2) \,A_1 \,(Z_1,Z_2)^\top +  
\exp \left((Z_3,Z_4, Z_5) \,A_2 \, (Z_3,Z_4, Z_5)^\top\right),
\end{equation*}
with matrices $A_1 \in \R^{2\times 2}$ and $A_2 \in \R^{3\times 3}$. Thus, the first part is a bilinear form in $(Z_1,Z_2)^\top$, and the second term is a general form involving the remaining components of $\bZ$. We compute the Shapley values $(\phi_j)_{j=1}^5$, the Paired-Sampling KernelSHAP $\widehat{\bphi}_n^{(PS)}$ for $n=100$ random samples, and the Paired-Sampling PermutationSHAP $(\widehat{\phi}^{(1)}_j)_{j=1}^5$ for one single paired permutation.

\begin{table}[htb]
\begin{center}
{\small
\begin{tabular}{|l||cc|ccc|}
\hline
& $j=1$& $j=2$ & $j=3$ & $j=4$ & $j=5$ \\\hline\hline
exact Shapley values $(\phi_j)_{j=1}^5$
&$-0.8153$ & $-0.1830$ & $-1.1116$ & $ 0.5998 $ & $0.7837$\\
PS KernelSHAP $\widehat{\bphi}_n^{(PS)}$
 &$-0.8593 $ & $-0.1890 $ & $-1.1031$ & $  0.7320 $ & $ 0.6929 $\\
PS PermutationSHAP $(\widehat{\phi}^{(1)}_j)_{j=1}^5$
&$ -0.8153 $ & $-0.1830 $ & $-1.5522$ & $  1.4810 $ & $ 0.3431$ \\
\hline
\end{tabular}}
\end{center}
\caption{Resulting Shapley values and their estimates (PS stands for Paired-Sampling).}
\label{table 1}
\end{table}
Table \ref{table 1} verifies that the Paired-Sampling PermutationSHAP gets the additive part (first two Shapley values for $j=1,2$) correct with one single paired permutation, where as the kernel version does not. The remaining Shapley values for $j=3,4,5$ can in both cases only be determined asymptotically (using the above asymptotic normality results). This closes the example.
\EndExample
\end{example}

The additive recovery property of the Paired-Sampling PermutationSHAP applies to any partition of the value function into disjoint additive parts. Assume that $({\cal A}_k)_{k=1}^K$ gives a partition of the grand coalition ${\cal Q}=\{1,\ldots, q\}$ with non-empty subsets ${\cal A}_k\neq \emptyset$ for all $1\le k \le K$. We define the feature subsets $\bZ_{{\cal A}_k}=(Z_j)_{j \in {\cal A}_k}$, for $1\le k \le K$, and we assume that the value function additively decomposes as
\begin{equation} \label{additive decomposition}
\nu(\bZ) = \sum_{k=1}^K \nu_k\left(\bZ_{{\cal A}_k}\right),
\end{equation}
for functions $\nu_k$ on the component subsets ${\cal A}_k \subset {\cal Q}$. We have the following proposition.
\begin{prop}\label{prop additive}
Assume the value function $\nu$ satisfies \eqref{additive decomposition}. Denote by
$(\widehat{\phi}^{(1)}_j)_{j=1}^q$ the Paired-Sampling PermutationSHAP values of an arbitrary permutation $\pi^{(1)}$. There is the additive recovery property
\begin{equation}\label{additive recovery permutation}
\sum_{j \in {\cal A}_k} \widehat{\phi}^{(1)}_j =
\sum_{j \in {\cal A}_k} \phi_j, \qquad \text{ for all $1\le k \le K$.}
\end{equation}
\end{prop}
\medskip
\begin{rems}\normalfont
\begin{itemize}
\item
The Paired-Sampling KernelSHAP does not satisfy any similar property, this can easily be verified by a numerical example. From this result we conclude that the permutation version has the advantage over the kernel version that it allocates the correct aggregated credits (Shapley value estimates) to all additive components 
$(\nu_k)_{k=1}^K$ of the value function $\nu$, and the asymptotic normality result is only needed to find the correct distribution within these additive components.
\item The additive recovery property \eqref{additive recovery permutation} holds for the PermutationSHAP even if we only consider a single permutation $\pi$ without its reverted version $\rho(\pi)$. Consequently, if we compute the PermutationSHAP estimates for different permutations $\pi$, and if for all these permutations we receive the same additive behavior \eqref{additive recovery permutation}, then we have identified the additive terms \eqref{additive decomposition} of the value function $\nu$.
\item These additive terms $(\nu_k)_{k=1}^K$ can also be found by the asymptotic covariance matrix $\Sigma$ \eqref{permutation Sigma}, showing uncorrelated blocks. That is, if the components of $\bZ$ are properly ordered, $\Sigma$ is a block-diagonal matrix.
\end{itemize}
\end{rems}
We provide an example that verifies these findings.

\begin{example}[additive recovery]\normalfont
We select the following value function. Choose $q=9$, $K=3$ and $\bZ_k=(Z_{3(k-1) +1} ,Z_{3(k-1) +2}, Z_{3(k-1) +3})^\top \in \{0,1\}^3$ for $1\le k \le K$. Then, we select the value function
\begin{equation*}
\nu(\bZ) = \sum_{k=1}^K 
\exp \left(\bZ_k^\top \,A_k \, \bZ_k\right),
\end{equation*}
with matrices $A_k \in \R^{3\times 3}$. Thus, there are $K=3$ additive parts; note that these three parts are not bilinear forms, so we do not expect to be able to recover the exact Shapley values $(\phi_j)_{j=1}^q$ by the paired-sampling versions. However, Proposition \ref{prop additive} tells us that we can find exactly the aggregated Shapley values within the additive groups. We verify this. Again, we select the matrices $A_k$ by i.i.d.~standard normal variables.
Then, we compute the Paired-Sampling PermutationSHAP estimates $\widehat{\phi}^{(1)}_j$ from one single (arbitrary) permutation, and we verify the additive recovery property
\eqref{additive recovery permutation}. For comparison, we also compute the Paired-Sampling KernelSHAP version for $n=100$
instances. We see that this KernelSHAP does not have the additive recovery property. The results are shown in Table \ref{table 2}, and they justify our statements.

\begin{table}[htb]
\begin{center}
{\small
\begin{tabular}{|l||ccc|}
\hline
& $k=1$& $k=2$ & $k=3$ \\\hline\hline
exact Shapley values
&$0.3981$ & $  3.5639 $ & $-0.7954$
\\
Paired-Sampling KernelSHAP
 &$0.7141 $ & $ 3.3924 $ & $-0.9400
 $\\
Paired-Sampling PermutationSHAP 
&$0.3981$ & $  3.5639 $ & $-0.7954$ \\
\hline
\end{tabular}}
\end{center}
\caption{Analysis of the additive recovery property \eqref{additive recovery permutation} for the three groups $({\cal A}_k)_{k=1}^3$.}
\label{table 2}
\end{table}

\begin{equation*}
\Sigma
= \begin{pmatrix}
1.85 &-0.92& -0.92 & 0.00 & 0.00 & 0.00 & 0.00 & 0.00&  0.00\\
 -0.92 & 1.85 &-0.92 & 0.00 & 0.00 & 0.00  &0.00&  0.00&  0.00\\
 -0.92 &-0.92 & 1.85 & 0.00 & 0.00 & 0.00 & 0.00 & 0.00 & 0.00\\
  0.00 & 0.00 & 0.00 & 3.11 &-1.56 &-1.56 & 0.00 & 0.00 & 0.00\\
  0.00 & 0.00 & 0.00 &-1.56 & 3.11 &-1.56 & 0.00 & 0.00  &0.00\\
  0.00 & 0.00 & 0.00 &-1.56 &-1.56 & 3.11 & 0.00 & 0.00 & 0.00\\
  0.00 & 0.00 & 0.00 & 0.00 & 0.00 & 0.00 & 0.03 &-0.02 &-0.02\\
  0.00 & 0.00 & 0.00 & 0.00 & 0.00 & 0.00 &-0.02 & 0.03 &-0.02\\
  0.00 & 0.00 & 0.00 & 0.00 & 0.00 & 0.00 &-0.02 &-0.02  &0.03
\end{pmatrix}
\end{equation*}
The above matrix shows the asymptotic covariance matrix $\Sigma$ of the limiting Gaussian distribution \eqref{permutation Sigma} of our example. We observe a block-diagonal structure, which fully identifies the additive decomposition $({\cal A}_k)_{k=1}^q$ of the grand coalition ${\cal Q}$. Of course, the application of this result in practice faces two difficulties. First, $\Sigma$ needs to be estimated from an observed sample, which results in noisy estimates that are not precisely zero, thus, we may need to manually set the entries to zero that are not significantly different from zero. Second, we may need to reorder the components of $\bZ$ to receive a block-diagonal matrix, i.e., typically the components of the feature $\bZ$ will have an arbitrary order. This closes the example.
\EndExample
\end{example}

\section{Conclusions}
\label{Section Conclude}
This paper describes important properties of exact KernelSHAP, exact PermutationSHAP, and their sampling versions. First, exact KernelSHAP provides the same results as exact PermutationSHAP, and these results are in line with the exact Shapley values. Second, paired-sampling versions coincide with their exact counterparts as long as the value function is a bilinear form, i.e., as long as it does not contain interactions of orders bigger than 2. Third, we provide asymptotic normality results for the sampling versions of the SHAP computations, and we verify them empirically. From these results we conclude that the sampling errors are smaller for the permutation version if compared on the level of samples $n$. However, since the two versions -- kernel and permutation -- involve different numbers of evaluations of  the value function, the preference may change to the kernel version if measured in terms of these numbers of evaluations.
Fourth, the paired-sampling PermutationSHAP poses the additive recovery property, whereas its kernel sampling version counterpart does not. This latter item is very important in practice to identify interaction clusters among the feature components, and this clearly gives support to using the paired-sampling PermutationSHAP version. Moreover, the covariance matrix of the asymptotic normality result allows one to inspect and identify such interaction clusters.

\medskip

{\small 
\renewcommand{\baselinestretch}{.51}
}

\newpage

\appendix

\section{Proofs}
{\Beweis
{\bf Proof of Proposition \ref{Prop2.3}.}
We compute the difference
\begin{equation*}
{\cal J}^{-1}{\cal I}{\cal J}^{-1}-{\cal J}_2^{-1}{\cal I}_2 {\cal J}_2^{-1}
= {\cal J}^{-1}\left({\cal I}-\frac{1}{4}{\cal I}_2\right) {\cal J}^{-1}.
\end{equation*}
This allows us to focus on the difference ${\cal I}-{\cal I}_2/4$. It is given by
\begin{eqnarray*}
{\cal I}-\frac{1}{4}{\cal I}_2&=&
\E_{\bZ \sim p}
\left[ \left(\nu(\bZ) -Z_q \nu(\bOne)- \left(\widetilde{\bZ}-Z_q\widetilde{\bOne}\right)^\top \widetilde{\bphi}\right)^2\left(\widetilde{\bZ}-Z_q\widetilde{\bOne}\right)\left(\widetilde{\bZ}-Z_q\widetilde{\bOne}\right)^\top\right]
\\&&-
\E_{\bZ \sim p}
\left[\left(\frac{\nu(\bZ)+\nu(\bOne)-\nu(\bOne-\bZ)}{2}-Z_q \nu(\bOne) - \left(\widetilde{\bZ}-Z_q\widetilde{\bOne}\right)^\top \widetilde{\bphi}\right)^2\left(\widetilde{\bZ}-Z_q\widetilde{\bOne}\right)\left(\widetilde{\bZ}-Z_q\widetilde{\bOne}\right)^\top\right].
\end{eqnarray*}
We consider the square bracket of the second term. Applying Jensen's inequality we have
\begin{eqnarray*}
&&\hspace{-1cm}
\left(\frac{\nu(\bZ)+\nu(\bOne)-\nu(\bOne-\bZ)}{2}-Z_q \nu(\bOne) - \left(\widetilde{\bZ}-Z_q\widetilde{\bOne}\right)^\top \widetilde{\bphi}\right)^2
\\&=&
\left(\frac{1}{2}\left(\nu(\bZ)-Z_q \nu(\bOne) - \left(\widetilde{\bZ}-Z_q\widetilde{\bOne}\right)^\top \widetilde{\bphi}
\right)
+
\frac{1}{2}\left(\nu(\bOne)-\nu(\bOne-\bZ)-Z_q \nu(\bOne) - \left(\widetilde{\bZ}-Z_q\widetilde{\bOne}\right)^\top \widetilde{\bphi}
\right)
\right)^2
\\&\le&
\frac{1}{2}\left(\nu(\bZ)-Z_q \nu(\bOne) - \left(\widetilde{\bZ}-Z_q\widetilde{\bOne}\right)^\top \widetilde{\bphi}
\right)^2
+
\frac{1}{2}\left(\nu(\bOne)-\nu(\bOne-\bZ)-Z_q \nu(\bOne) - \left(\widetilde{\bZ}-Z_q\widetilde{\bOne}\right)^\top \widetilde{\bphi}
\right)^2
\\&=&
\frac{1}{2}\left(\nu(\bZ)-Z_q \nu(\bOne) - \left(\widetilde{\bZ}-Z_q\widetilde{\bOne}\right)^\top \widetilde{\bphi}
\right)^2
+
\frac{1}{2}\left(\nu(\bZ')-Z'_q \nu(\bOne) - \left(\widetilde{\bZ}'-Z'_q\widetilde{\bOne}\right)^\top \widetilde{\bphi}
\right)^2.
\end{eqnarray*}
This completes the proof.
\EndProof}

\bigskip

{\Beweis
{\bf Proof of Proposition \ref{bilinear SHAP}.}
To prove Proposition \ref{bilinear SHAP} we come back to the permutation SHAP formula \eqref{permutation SHAP}. Sample a random permutation $\pi$ of the ordered set $(1,\ldots, q)$. The contribution of this permutation $\pi$ to the
Shapley value $\phi_j$ under \eqref{binary interactions} is
\begin{equation*}
\nu\left({\cal C}_{\pi, j}\cup \{j\}\right)
-\nu\left({\cal C}_{\pi, j}\right) 
= \frac{1}{q!}\left(a_{j,j} + \sum_{k \in {\cal C}_{\pi,j}} \left(a_{j,k} + a_{k,j}\right)\right).
\end{equation*}
At the same time we consider the reverse of permutation $\rho(\pi)$ of $\pi$, and we have
\begin{equation*}
{\cal C}_{\rho(\pi), j} = \left\{ \pi_{\kappa(j)+1}, \ldots, \pi_q  \right\}={\cal Q}\setminus 
\left({\cal C}_{\pi, j} \cup \{j\}\right),
\end{equation*}
where $\pi_{\kappa(j)}=j$; see \eqref{kappa function}.
The contribution of this reverted permutation to the
Shapley value $\phi_j$ is
\begin{equation*}
\nu\left({\cal C}_{\rho(\pi), j}\cup \{j\}\right)
-\nu\left({\cal C}_{\rho(\pi), j}\right) 
=\frac{1}{q!}\left( a_{j,j} + \sum_{k \in {\cal Q}\setminus ({\cal C}_{\pi, j} \cup \{j\})} \left(a_{j,k} + a_{k,j}\right)\right).
\end{equation*}
From this we observe that aggregating the contributions of 
$\pi$ and $\rho(\pi)$ results in the bilinear form case
\eqref{binary interactions} in
\begin{equation*}
\left(\nu\left({\cal C}_{\pi, j}\cup \{j\}\right)
-\nu\left({\cal C}_{\pi, j}\right) \right)
+\left(\nu\left({\cal C}_{\rho(\pi), j}\cup \{j\}\right)
-\nu\left({\cal C}_{\rho(\pi), j}\right)\right) 
=
\frac{1}{q!}\sum_{k=1}^q \left(a_{j,k} + a_{k,j}\right).
\end{equation*}
The right-hand side is independent of $\pi$, and since we have $q!/2$ permutations $\pi$ with reversed pairs $\rho(\pi)$ completes the proof.
\EndProof
}

\bigskip

{\Beweis
{\bf Proof of Proposition \ref{bilinear SHAP 2}.}
The proof is quite similar to the one of 
Proposition \ref{bilinear SHAP}. For $j \in \{1,\ldots, d\}$, we have
contribution of permutation $\pi$ to the
Shapley value $\phi_j$
\begin{equation*}
\nu\left({\cal C}_{\pi, j}\cup \{j\}\right)
-\nu\left({\cal C}_{\pi, j}\right) 
= \frac{1}{q!}\left(a_{j,j} + \sum_{k \in {\cal C}_{\pi,j} \cap \{1,\ldots, d\}} \left(a_{j,k} + a_{k,j}\right)\right).
\end{equation*}
The main difference to the previous proof is the intersection with $\{1,\ldots, d\}$ in the summation. This is obtained because the first $d$ components of $\bZ$ do not interact with the last $q-d+1$ one. The remainder of the proof is then rather similar to the one of Proposition \ref{bilinear SHAP}, and the last step is verified by comparing the contribution of the permutation $\pi$ and its reverted version $\rho(\pi)$ to Corollary \ref{cor xz}. This completes the proof.
\EndProof}

\bigskip

{\Beweis
{\bf Proof of Proposition \ref{prop additive}.}
The additive decomposition \eqref{additive decomposition} given by
\begin{equation*} 
\nu(\bZ) = \sum_{k=1}^K \nu_k\left(\bZ_{{\cal A}_k}\right),
\end{equation*}
leads to $K$ cooperative games $\nu_k$, where in game $k$ only the players $\bZ_{{\cal A}_k}$ play, and the remaining players are dummy players. Thus, based on the linearity axiom (A4) and the dummy player axiom (A3), the credit $\phi_j$ received by player $j\in {\cal A}_k$ is fully determined by game $k$. Consequently, ${\cal A}_k$ forms the simplified grand coalition for game $k$, and $\nu_k({\cal A}_k)$ describes the whole payoff shared by these players (note that $\nu_k({\cal A}_k)$ uses a slight abuse of notation, but we refrain from giving the fully precise definition as its meaning is pretty clear).\\
We now consider the Shapley value estimate from one single permutation $\pi$. It is for $j \in {\cal A}_k$ given by
\begin{equation*}
\nu({\cal C}_{\pi, j} \cup \{j\}) - \nu({\cal C}_{\pi, j})
=
\nu_k\left(({\cal C}_{\pi, j}\cap {\cal A}_k) \cup \{j\}\right) - \nu_k\left({\cal C}_{\pi, j}\cap {\cal A}_k\right),
\end{equation*}
all the components not belonging to ${\cal A}_k \ni j$ cancel because of the additive decomposition \eqref{additive decomposition}. We denote by $\pi_{{\cal A}_k}$ the selected permutation $\pi$, but only restricted to the components in ${\cal A}_k$. This allows us to rewrite the previous formula as
\begin{equation*}
\nu({\cal C}_{\pi, j} \cup \{j\}) - \nu({\cal C}_{\pi, j})
=
\nu_k\left({\cal C}_{\pi_{{\cal A}_k}, j}\cup \{j\}\right) - \nu_k\left({\cal C}_{\pi_{{\cal A}_k}, j}\right).
\end{equation*}
If we now sum these over all components $j\in {\cal A}_k$ we receive a telescoping sum
\begin{equation*}
\sum_{j \in {\cal A}_k}
\nu({\cal C}_{\pi, j} \cup \{j\}) - \nu({\cal C}_{\pi, j})
=\sum_{j \in {\cal A}_k}
\nu_k\left({\cal C}_{\pi_{{\cal A}_k}, j}\cup \{j\}\right) - \nu_k\left({\cal C}_{\pi_{{\cal A}_k}, j}\right)
=\nu_k\left({\cal A}_k\right) - \nu_k\left(\emptyset\right)
=\sum_{j \in {\cal A}_k} \phi_j,
\end{equation*}
where the last identity follows from the efficiency axiom (A1) for cooperative game $\nu_k$. This proves that every single permutation $\pi$ provides the additive recovery \eqref{additive recovery permutation}, and so will the average over the permutation $\pi$ and its paired one $\rho(\pi)$. This completes the proof.
\EndProof}

\end{document}